\pdfoutput=1

\documentclass[11pt]{article}

\usepackage[]{emnlp}

\usepackage{times}
\usepackage{latexsym}

\usepackage[T1]{fontenc}

\usepackage[utf8]{inputenc}

\usepackage{microtype}

\usepackage{inconsolata}

\usepackage{header}
\title{Evaluating the Impact of Model Scale for \\ Compositional Generalization in Semantic Parsing}

\author{
Linlu Qiu$^1$\thanks{\enskip Work done as part of the Google AI Residency program.} \quad 
Peter Shaw$^2$ \quad 
Panupong Pasupat$^2$ \quad 
Tianze Shi$^2$ \quad \\ 
{\bf Jonathan Herzig}$^2$ \quad
{\bf Emily Pitler}$^2$ \quad
{\bf Fei Sha}$^2$ \quad
{\bf Kristina Toutanova}$^2$ \\[.5em]
$^1$Massachusetts Institute of Technology \quad $^2$Google Research \\
\small{\texttt{linluqiu@mit.edu,\{petershaw,ppasupat,tianze,jherzig,epitler,fsha,kristout\}@google.com}}
}

\begin{document}
\maketitle
\begin{abstract}
Despite their strong performance on many tasks, pre-trained language models have been shown to struggle on out-of-distribution compositional generalization. Meanwhile, recent work has shown considerable improvements on many NLP tasks from model scaling. Can scaling up model size also improve compositional generalization in semantic parsing? We evaluate encoder-decoder models up to 11B parameters and decoder-only models up to 540B parameters, and compare model scaling curves for three different methods for applying a pre-trained language model to a new task: fine-tuning all parameters, prompt tuning, and in-context learning. We observe that fine-tuning generally has flat or negative scaling curves on out-of-distribution compositional generalization in semantic parsing evaluations. In-context learning has positive scaling curves, but is generally outperformed by much smaller fine-tuned models. Prompt-tuning can outperform fine-tuning, suggesting further potential improvements from scaling as it exhibits a more positive scaling curve. Additionally, we identify several error trends that vary with model scale. For example, larger models are generally better at modeling the syntax of the output space, but are also more prone to certain types of overfitting. Overall, our study highlights limitations of current techniques for effectively leveraging model scale for compositional generalization, while our analysis also suggests promising directions for future work.

\end{abstract}
\section{Introduction}

Compositional generalization is the ability to generalize to novel combinations of previously observed elements. For example, we may ask a model to interpret ``she loves the dog'' when ``she'', ``loves'', and ``the dog'' were seen separately but not in combination with each other during training. Improving compositional generalization is believed to be important for approaching human-like language understanding~\cite{lake2017building, battaglia2018relational}.  In addition, models that are deployed for real-world applications often need to generalize to new compositions of elements not well-represented in static and often biased annotated training sets \cite{herzig2019don, yin2021compositional}. In this paper we focus on compositional generalization for semantic parsing, the task of mapping utterances to logical forms with precisely defined semantics.

Despite their strong performance on many tasks, pre-trained language models\footnote{We use the term ``language model'' (LM) broadly to refer to models based on generic encoder-decoder or decoder-only architectures that are pre-trained primarily using masked or autoregressive language modeling objectives, such as T5~\cite{raffel2019exploring}, BART~\cite{lewis-etal-2020-bart}, GPT-3~\cite{brown2020language}, and PaLM~\cite{chowdhery2022palm}.} (LMs) such as T5~\cite{raffel2019exploring} have been shown to struggle on compositional generalization~\cite{lake2018generalization, furrer2020compositional, shaw-etal-2021-compositional}.
However, recent work has shown considerable improvements across a range of NLP tasks from scaling up model size~\cite{brown2020language, chowdhery2022palm}. \emph{Can scaling up the number of parameters of pre-trained language models also improve compositional generalization in semantic parsing?}

Understanding the relationship between model size and compositional generalization ability has important implications for future work. If increasing model size does not improve compositional generalization in semantic parsing, this would run counter to many scaling trends in NLP, and highlight a potential limitation of advances that could be expected from scaling alone.  On the other hand, if gains from scale are very strong, larger models pre-trained on more and higher quality unlabeled data would point to a successful (albeit expensive) alternative to current work that has focused on developing specialized architectures and other novel methods (see \S\ref{sec:related_work} for a brief survey).

This naturally raises a second question: \emph{Does scaling behavior for compositional generalization in semantic parsing depend on the method of applying pre-trained language models?} Full fine-tuning of model parameters is a standard approach for applying LMs to end tasks, and T5 performance with fine-tuning has been measured for compositional generalization up to the scale of 11 billion parameters~\cite{shaw-etal-2021-compositional, furrer2020compositional}. More recently, variants of prompting with or without some parameter tuning have become commonly used as well (\newcite{ppp_survey_neubig} provides a comprehensive survey). Although there are studies on large models for semantic parsing with such methods, e.g. with in-context learning using GPT-3~\cite{brown2020language,shin-etal-2021-constrained, shin2021few, rubin-etal-2022-learning, rajkumar2022evaluating}, they do not focus on compositional generalization.

In this paper, we offer the first systematic study of scaling curves measuring compositional generalization in semantic parsing versus model size for LMs under multiple task adaptation techniques.\footnote{In this paper we use ``task adaptation'' to refer to application of a pre-trained LM to a downstream task.} %
We focus on a set of compositional semantic parsing challenges and evaluate model sizes up to 540 billion parameters. We compare scaling curves for an encoder-decoder model (T5)~\cite{raffel2019exploring} and a decoder-only model (PaLM)~\cite{chowdhery2022palm}. We measure the impact of scale for three different task adaptation methods, representing the spectrum of tuning all of the model's parameters for the end task (full fine-tuning) to none of the parameters (in-context learning). In addition to these two ends of the spectrum, we choose prompt tuning~\cite{lester-etal-2021-power} as a representative of parameter-efficient task adaptation methods~\cite{he2022towards}.

We identify several error trends that change as a function of model size and task adaptation technique. Additionally, we analyze how different types of errors, distribution shift, output representations, and different retrievers for constructing prompts for in-context learning affect scaling trends.
The key observations of our study can be summarized as follows:
\begin{itemize}
 \item When fine-tuning LMs, we generally observe flat or negative scaling curves for compositional generalization in semantic parsing.
 \item Prompt tuning can outperform standard fine-tuning for larger models, as it exhibits a more positive scaling curve. This suggests the potential for further improvements from scaling combined with prompt-tuning or potentially other parameter-efficient methods for task transfer.
 \item We observe positive scaling curves for in-context learning, but performance for the largest model size is generally worse than fine-tuning performance for much smaller models. 
 \item We observe both positive and negative trends for different types of errors as a function of model size and task adaptation technique. For example, larger models perform better at modeling the syntax of the output space, but can also be more prone to certain types of overfitting, especially when fine-tuned.
\end{itemize}

\section{Related Work}
\label{sec:related_work}

\paragraph{Compositional Generalization}
Many approaches have been proposed to improve compositional generalization in semantic parsing, including compositional data augmentation~\cite{jia2016data, andreas-2020-good, Akyrek2021LearningTR, Oren2021FindingNI, qiu2022improving, yang2022subs}, specialized architectures~\cite{li2019compositional, russin2019compositional, gordon2020permutation, liu2020compositional, nye2020learning, chen2020compositional, zheng2020compositional, oren2020improving, herzig2020span, ruiz2021iterative, wang2021structured}, ensemble models~\cite{shaw-etal-2021-compositional}, different Transformer variations~\cite{csordas2021devil, ontanon2021making}, intermediate representations~\cite{herzig2021unlocking, shin-etal-2021-constrained}, meta-learning~\cite{lake2019compositional, conklin2021meta, zhu2021learning}, and auxiliary objectives to bias attention in encoder-decoder models~\cite{yin2021compositional, jiang2021inducing}. \citet{furrer2020compositional} compare pre-trained models with specialized architectures, but they focus on evaluating the impact of pre-training and only fine-tune encoder-decoder models up to 11B. \citet{tsarkov2021cfq} evaluate the impact of training size, but keep the computational cost fixed. 

\paragraph{Scaling}

Many existing studies investigate the scaling of neural networks to better understand scaling laws and optimize training budget~\cite{hestness2017deep, kaplan2020scaling, bornschein2020small, ghorbani2021scaling, bahri2021explaining, tay2021scale, rae2021scaling, hoffmann2022training,ivgi2022scaling}. The scaling behavior of many tasks have been shown to be predictable and generalization error generally decreases with scale~\cite{geiger2019scaling, rosenfeld2020constructive, henighan2020scaling}, with some exceptions showing the limits of large scale pre-training~\cite{abnar2021exploring}. 
\citet{hern2021scaling} study scaling laws for transfer and find benefits of pre-training. 

\paragraph{Task Adaptation}
With the advances in capabilities of pre-trained LMs, a large set of techniques for transferring or adapting these models to end tasks of interest have been developed~\cite{wang2022what}.  \emph{Fine-tuning} all or most model parameters for each end task has been the standard approach for encoder-only models of the size of BERT~\cite{devlin2018bert} and RoBERTa~\cite{liu2019roberta} and encoder-decoder models like T5~\cite{raffel2019exploring}. Recently, variants of prompting, which uses a language model to directly make end-task predictions, have become a popular paradigm to adapt models to new tasks~\cite{ppp_survey_neubig}. \emph{In-context learning} shows the ability of LMs to learn to perform a novel task only by a small number of demonstrations during inference~\cite{brown2020language}. \emph{Prompt tuning}~\cite{lester-etal-2021-power, li-liang-2021-prefix, liu2021ptuning} learns a small number of parameters conditioning on frozen LMs. Many of these approaches can be seen as variants of parameter-efficient task transfer~\cite{he2022towards} and our selection of task adaptation methods covers representatives from the full spectrum of tuning none to all of a model's parameters for the end tasks. \newcite{schucher-etal-2022-power} investigate prompt tuning for semantic parsing. \newcite{wortsman2021robust} and \newcite{kumar2022finetuning} study task adaptation techniques to improve out-of-distribution generalization on other tasks, but not semantic parsing. \newcite{xie2022unifiedskg} propose the UnifiedSKG framework to leverage LMs for new structured knowledge grounding tasks including semantic parsing, but do not focus on compositional generalization.

\section{Experimental Setup}
\label{sec:experiments}

\subsection{Datasets}
We evaluate exact match accuracy on semantic parsing tasks where natural language utterances are mapped to meaning representations. We use both synthetic (COGS and CFQ) and non-synthetic datasets (GeoQuery and SMCalFlow-CS). More details on all datasets and splits are in Appendix~\ref{sec:appendix_dataset}.

\begin{figure}[ht!]
\centering
\includegraphics[width=\linewidth]{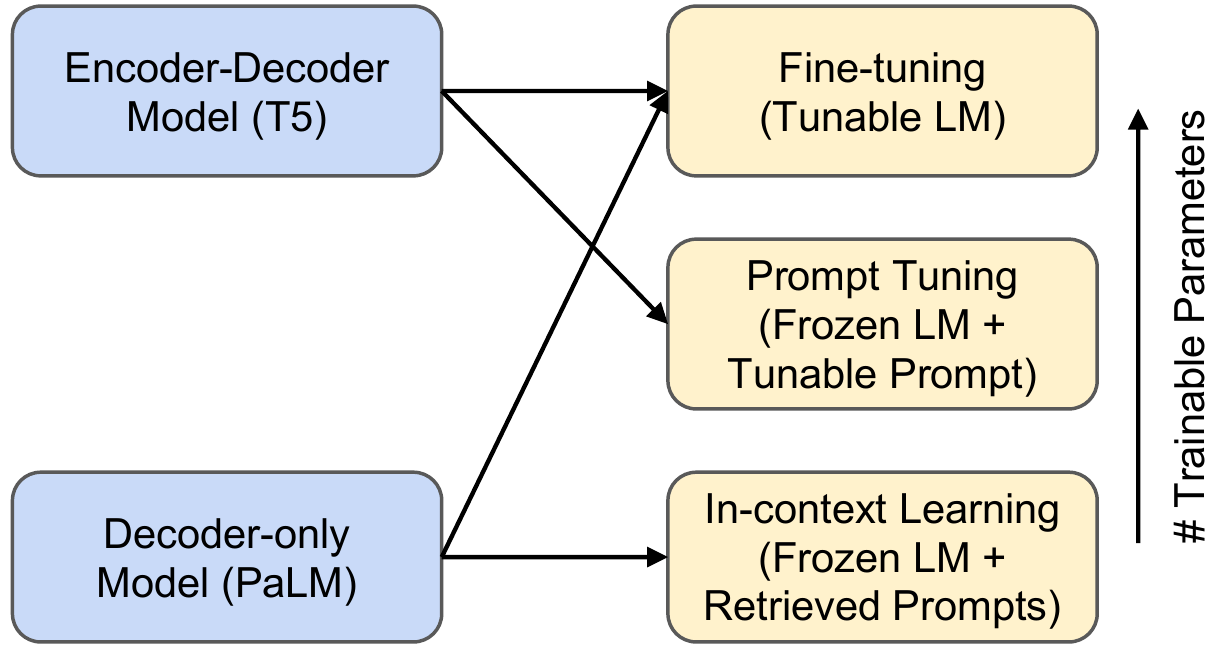}
\caption{Our experimental setup.}
\label{fig:setup}
\end{figure}

\paragraph{COGS}
The COGS dataset \cite{kim-linzen-2020-cogs} contains sentences paired with logical forms.
We use the in-distribution test set and generalization test set that tests generalization to novel linguistic structures.
We evaluate on a small subset with 50 examples from each test category (1050 examples total) due to computational constraints.
The main experiments convert the original lambda calculus outputs to equivalent variable-free forms~\cite{qiu2022improving}.
(\S\ref{sec:output_space} discusses how the output format affects the model.)

\paragraph{CFQ}
The CFQ dataset \cite{keysers2019measuring} contains questions paired with SPARQL queries. We use the random split and three Maximum Compound Divergence (MCD) splits from the original source. We evaluate on a subset with randomly sampled 1000 examples for each split. 

\begin{figure*}[t!]
\centering
\includegraphics[width=\linewidth]{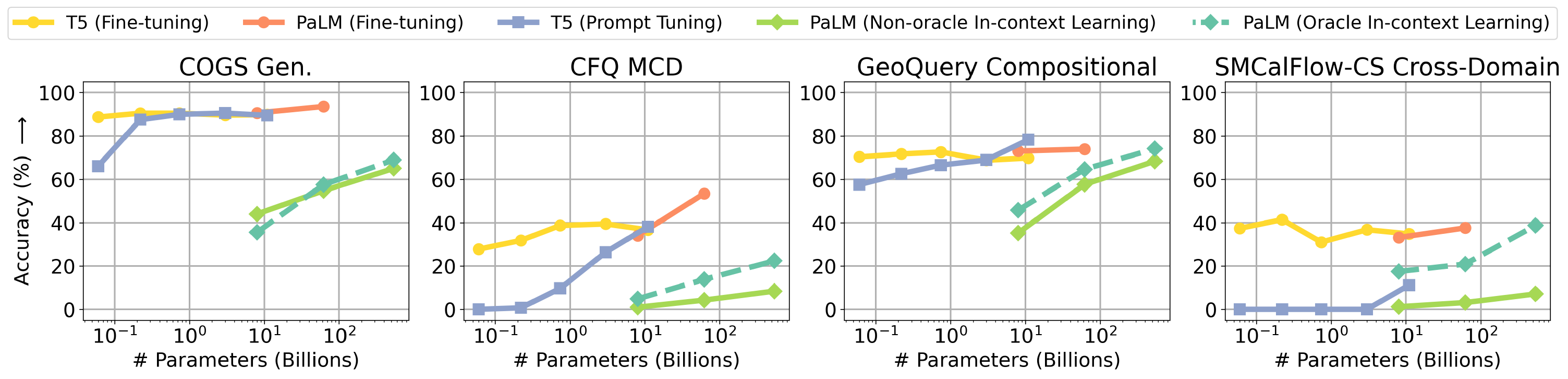}
\caption{Aggregated scaling curves for compositional splits of different datasets. Note that in-context learning with an oracle retriever (dashed) cannot be compared directly with other methods as it has access to the gold output.}
\label{fig:scale}
\end{figure*}

\paragraph{GeoQuery}
GeoQuery~\cite{zelle1996learning,tang2001using} contains human-authored questions paired with meaning representations. We report results on the standard data split as well as three compositional splits based on those introduced
in \citet{shaw-etal-2021-compositional}:
(1) the \emph{template} split, where abstract output templates in training and test data are disjoint~\cite{finegan2018improving};
(2) the \emph{TMCD} split, which makes the distributions of compounds in training and test data as divergent as possible;
and (3) the \emph{length} split.

\paragraph{SMCalFlow-CS}
SMCalFlow-CS is a \emph{compositional skills} split of SMCalFlow~\cite{andreas-etal-2020-task} proposed by \citet{yin2021compositional}. It contains single-turn sentences involving skills related to event creation and organization structure. We use LISPRESS~\cite{platanios-etal-2021-value}, which is a LISP-like serialization format, for programs.\footnote{Since the original SMCalFlow-CS release uses LISP format, which is extensively verbose and less suitable for neural seq2seq model, we re-ran the data generation pipeline to create new SMCalFlow-CS splits with LISPRESS format using \url{https://github.com/microsoft/compositional-generalization-span-level-attention}.}
The single-domain (S) test set has examples from a single domain, while the cross-domain (C) test set evaluates on examples that feature compositional skills  (e.g., ``create an event with my manager''). The evaluation considers a few-shot compositional learning scenario, where only a small number of cross-domain examples (8, 16, or 32) are seen during training. Additionally, we create a length split by using the longest 720 single-domain examples for evaluation and the rest single-domain examples for training.

\subsection{Models}

We study representatives of two classes of models: the encoder-decoder model T5~\cite{raffel2019exploring} and the decoder-only model PaLM~\cite{chowdhery2022palm}. We consider five T5 models with sizes ranging from 60M to 11B parameters, and three PaLM models with sizes ranging from 8B to 540B parameters.

We show our setup in Figure~\ref{fig:setup}. We evaluate three task adaptation techniques ranging from full to zero parameter updates: fine-tuning, prompt tuning, and in-context learning. We use T5 models (small to 11B) and PaLM models (8B and 62B) for fine-tuning. We perform prompt tuning with T5 models (small to 11B) following~\citet{lester-etal-2021-power}, but not PaLM as prompt tuning with decoder-only models has not been widely explored.\footnote{We leave evaluating decoder-only models with other parameter-efficient tuning methods such as prefix tuning~\cite{li-liang-2021-prefix} as future work.} %
We follow \citet{brown2020language} and use the decoder-only model PaLM (8B, 62B, and 540B) for in-context learning.\footnote{We did not use T5 models for in-context learning as they are trained using span corruption objectives. We also conducted preliminary in-context learning experiments using LM-adapted T5 models~\cite{lester-etal-2021-power}, but found the performance was much worse than using PaLM.} Further details on experiments are in Appendix~\ref{sec:appendix_experiment}.

\subsection{Retrievers}
\label{sec:experiment_retriever}

For in-context learning, we retrieve a small number of exemplars to construct prompts. We consider various formulations for unsupervised non-oracle retrievers (that have access to the input query only) and oracle retrievers (that have access to both the input and target output). We include oracle retrievers to approximate an upper bound on in-context learning performance. We include more retriever analysis in \S\ref{sec:retriever_analysis} and Appendix~\ref{sec:appendix_retriever_analysis}.

\paragraph{Non-oracle} We use the classical retrieval method BM25~\cite{robertson2009probabilistic} to retrieve the most similar exemplars for each test query. We also consider a BERT retriever that uses BERT-base~\cite{devlin2018bert} to encode the query and retrieve exemplars based on the cosine similarity of the [CLS] embeddings.

\paragraph{Oracle} We use BM25 similarity to the gold output instead of the input query following~\citet{rubin-etal-2022-learning}. We also consider target overlap retriever which retrieves exemplars based on the Jaccard similarity between \emph{compounds} in the gold output and those of the outputs in the training examples. We define compounds as combinations of parent and child symbols in the output, similarly to~\citet{shaw-etal-2021-compositional}. Additionally, we ensure exemplars contain all component symbols of the gold output, if they exist in the training set.

\section{Results \& Analysis}

\subsection{Main Results}
We show the aggregated scaling curves for different splits and datasets in Figure~\ref{fig:scale} (see Appendix~\ref{sec:individual-split-results} for results on individual splits).
We report the best in-context learning results of non-oracle and oracle retrievers using the maximum number of exemplars. We include ablations of retrievers in Appendix~\ref{sec:appendix_retriever_analysis} and other experiment details in Appendix~\ref{sec:appendix_experiment}.

\paragraph{}First, we generally observe flat or negative scaling curves when fine-tuning LMs except on the CFQ dataset, suggesting scaling with full fine-tuning is unlikely to be an effective solution for compositional generalization in semantic parsing as observed in \citet{shaw-etal-2021-compositional}, \citet{herzig2021unlocking}, and \citet{furrer2020compositional}.
\paragraph{}Second, scaling consistently improves in-context learning, but its performance is worse than that of a smaller fine-tuned model on the majority of the splits. The in-context learning performance is also highly dependent on the retriever, especially for splits with large training sets.

\paragraph{}Finally, the scaling curves for prompt tuning are more positive than the ones for fine-tuning, and prompt tuning sometimes outperforms fine-tuning for the same model size. %
This suggests scaling up models with parameter-efficient tuning methods could potentially further improve the compositional generalization ability of LMs.

\subsection{Error Analysis}
\label{sec:error_analysis}
In this section we analyze what types of errors models are making when they generate incorrect predictions, and how error trends change as a function of task adaptation technique and model size. We focus on non-synthetic datasets as they better represent the open problem of approaching human-like language understanding in practical scenarios.

\begin{figure}[t]
\centering
\includegraphics[width=\linewidth]{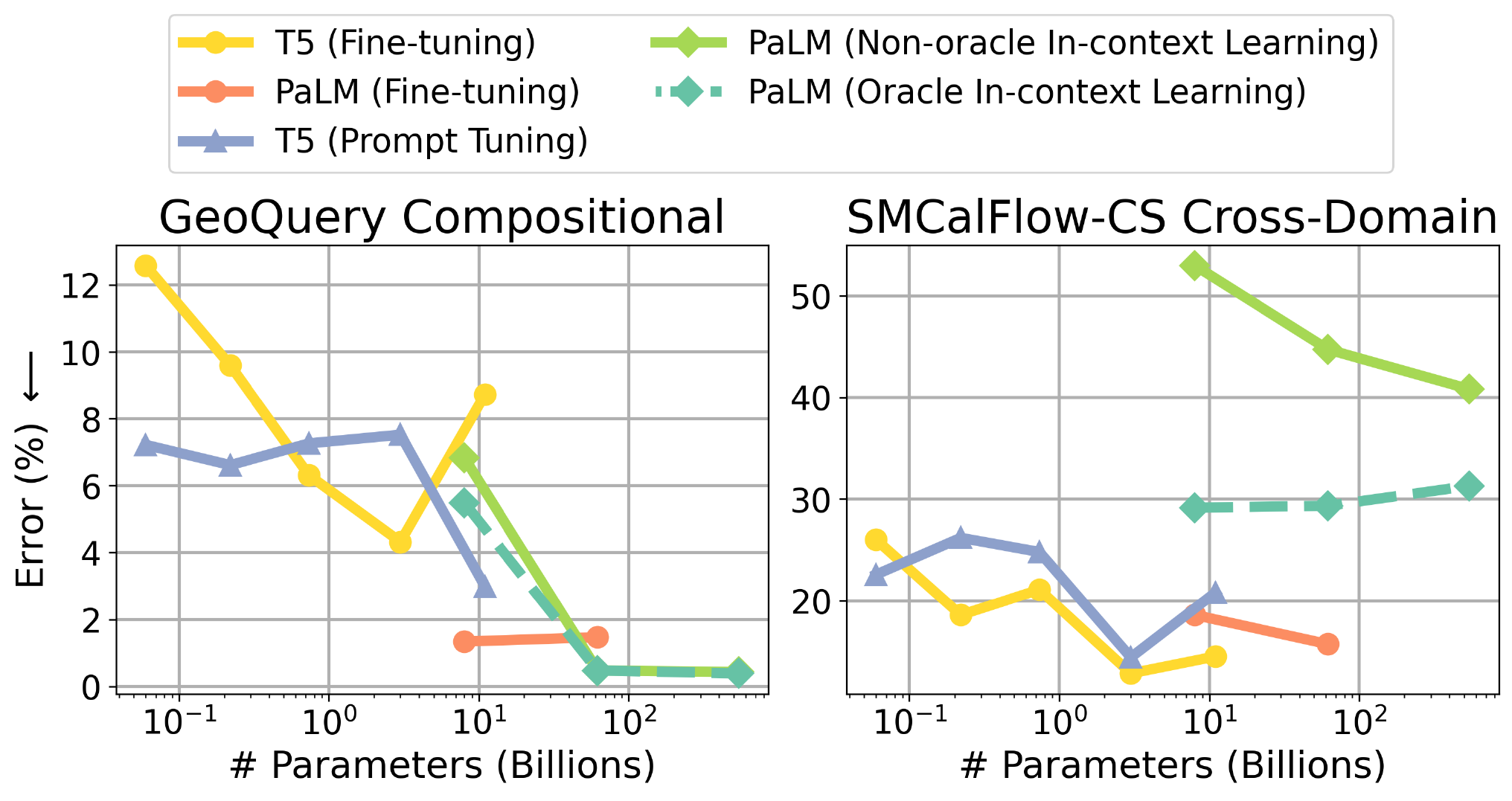}
\caption{Percentage of predictions that contain unbalanced parentheses, as an estimate of syntax errors.}
\label{fig:syntax_error}
\end{figure}

\paragraph{Syntax Errors} 
As we use unconstrained greedy decoding during inference, the generated prediction is not guaranteed to be syntactically valid. We measure the percentage of predictions that have unbalanced parentheses as an approximation to the overall rate of syntax errors. We aggregate the results from the different compositional splits for GeoQuery and SMCalFlow-CS (see Appendix~\ref{sec:individual-split-results} for error trends of individual splits).  The results are shown in Figure~\ref{fig:syntax_error}.
For the majority of splits and task adaptation methods, the number of syntax errors generally decreases when model scale grows, as observed in \citet{austin2021program}.
However, a large number of predictions are still syntactically incorrect, especially on SMCalFlow-CS, suggesting using constrained decoding to prevent generating invalid outputs can be an effective solution to improve performance in semantic parsing tasks~\cite{shin-etal-2021-constrained, shin2021few, scholak-etal-2021-picard}.

\paragraph{Compositional Errors}

\begin{figure}[t]
\centering
\includegraphics[width=\linewidth]{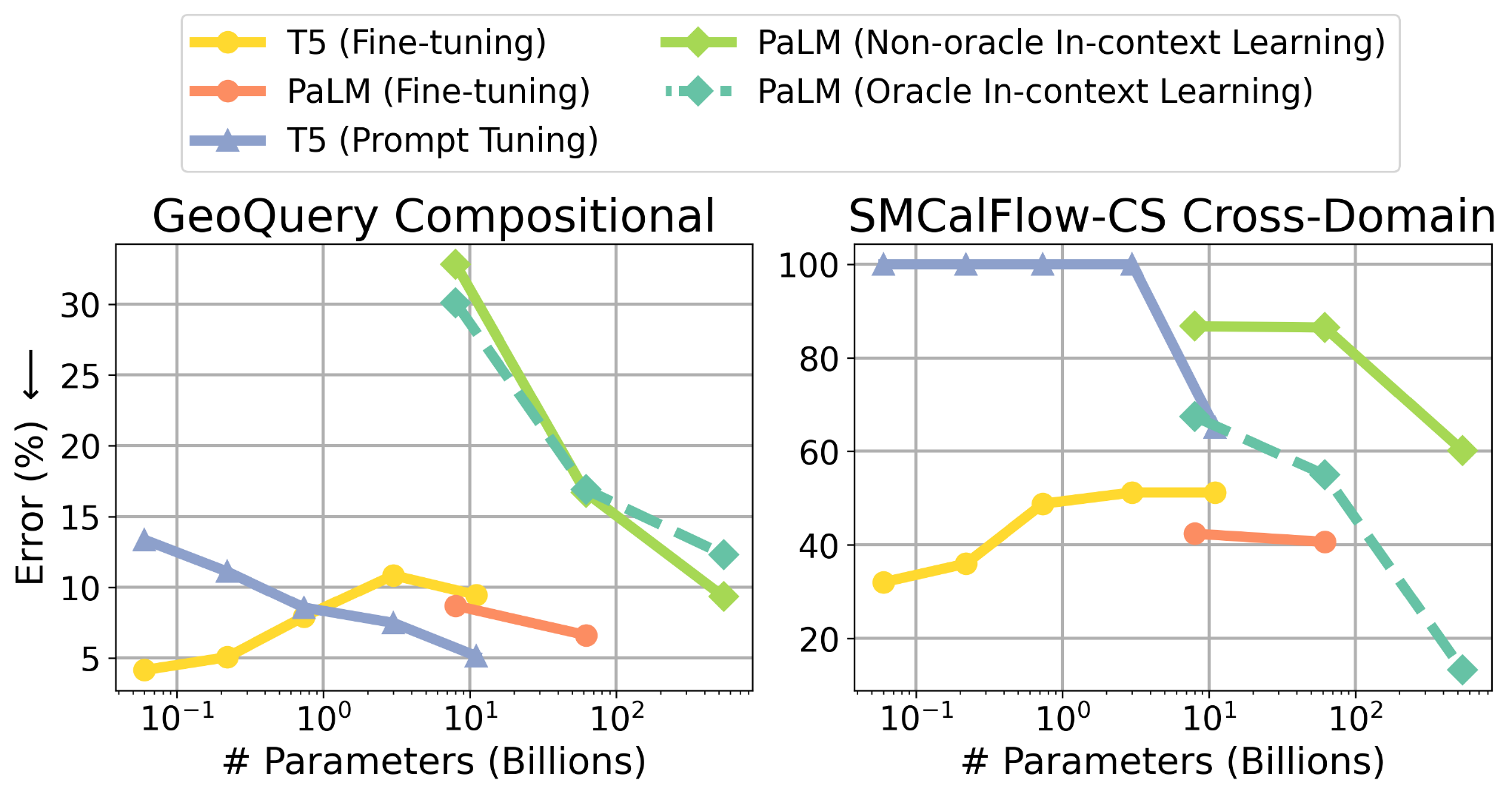}
\caption{Percentage of predictions where the output exactly matches an output seen in the training set on GeoQuery compositional splits (left). Percentage of predictions where the output only contains functions from a single domain on SMCalFlow-CS cross-domain splits (right).}
\label{fig:compositional_error}
\end{figure}

To evaluate the models' ability to recombine seen elements, we consider two measures. For GeoQuery compositional splits, since the entity names are anonymized, we measure the percentage of predictions where an output exactly matches an output seen in the training set and therefore does not include any recombination, leading to an error. For SMCalFlow-CS cross-domain splits, we investigate the errors involving the failure to recombine knowledge from the two domains. Specifically, we compute the percentage of predictions 
that only include functions from a single domain
and are therefore incorrect. Figure~\ref{fig:compositional_error} shows the results.
With fine-tuning, larger models are more likely to overfit to the training distribution and fail to recombine correctly.
In SMCalFlow-CS for example, larger models tend to generate single-domain predictions, while the information from the other domain is parsed as string literals; e.g., with input ``Please schedule Tuesday morning meeting with my team'', T5-small correctly predicts ``\texttt{\small (Event.attendees\_? (AttendeeListHasPeople (FindTeamOf (toRecipient (CurrentUser)))))}'' for the cross-domain part ``with my team'', while T5-11B tries to fit the information into the calendar event domain and outputs ``\texttt{\small (Event.subject\_? (?= "Team Meeting"))}''. 
However, we do not observe similar error trends for prompt tuning and in-context learning where limited or no parameters are updated.

\begin{figure}[t]
\centering
\includegraphics[width=\linewidth]{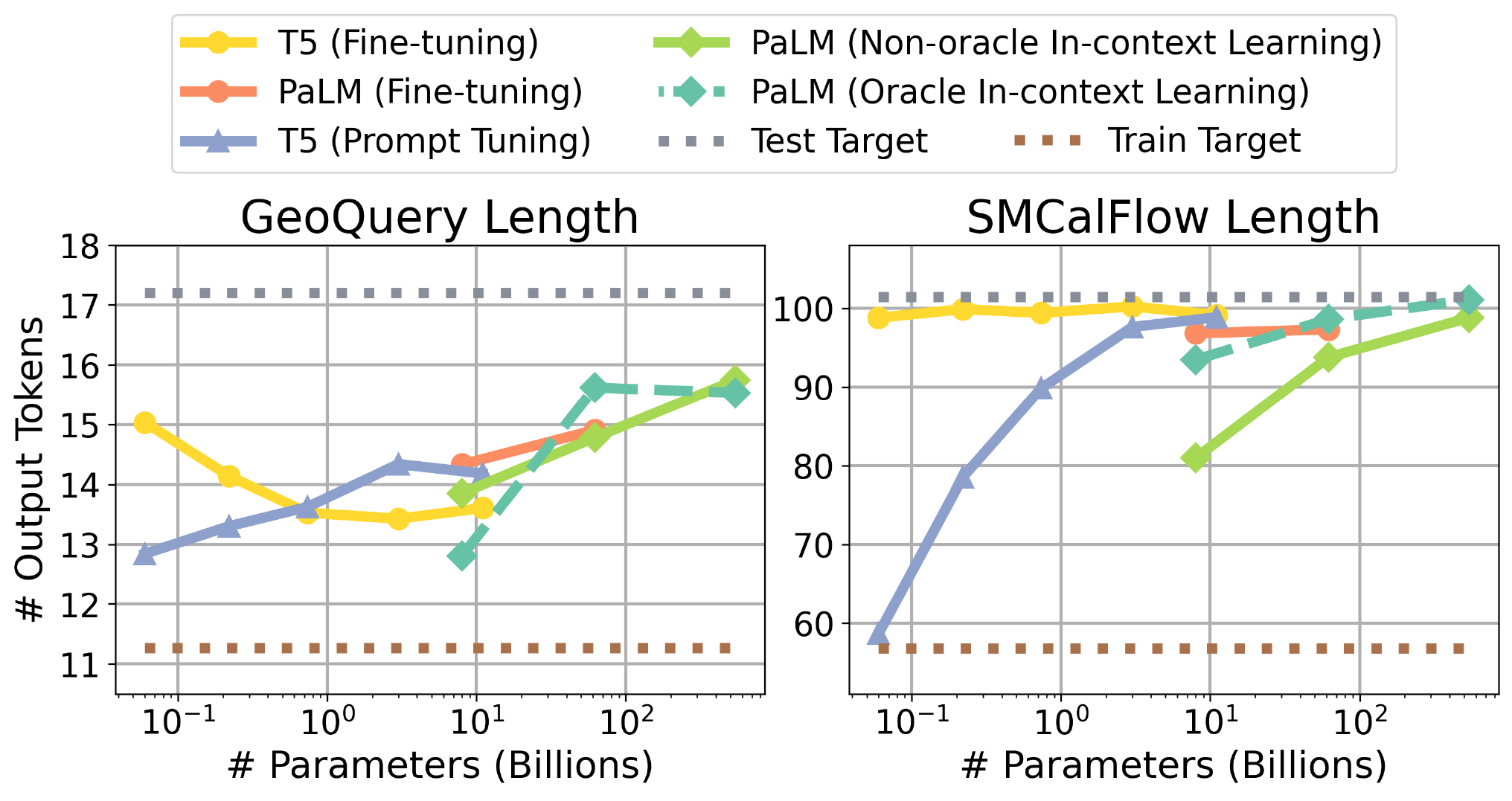}
\caption{Average number of tokens in the prediction. The dotted brown and black lines show the average target lengths in the training and test set, respectively.}
\label{fig:length}
\end{figure}

\paragraph{Length Extrapolation}
We compute the mean length of predictions for two length splits and show the results in Figure~\ref{fig:length}. The average length trends strongly correlate with the model performance trends. Model predictions are on average shorter than the gold outputs in the test set, suggesting all models have difficulty generating sequences longer than those seen during training~\cite{newman2020eos}. When using fine-tuning, the average length of T5 predictions decreases when the model becomes larger for the GeoQuery length split, but is roughly flat for the SMCalFlow length split. 
However, the average length of predictions increases when scaling up the PaLM model for in-context learning.
Scaling also increases the average length when using prompt tuning as opposed to fine-tuning for the T5 model, showing the potential of improving length extrapolation with methods like prompt tuning. 

\paragraph{Overfitting to Prior Output Distribution} Related to the compositional errors discussed above, one hypothesis is that fine-tuned language models overfit to and rely excessively on correlations present in the prior, input-independent distribution over outputs in the training data. For example, T5-11B fine-tuned on the GeoQuery TMCD1 split predicts
``\texttt{\small answer ( intersection ( river , 
\underline{loc$\_$2} ( m0 ) ) )}'' instead of ``\texttt{\small answer ( intersection ( river , 
\underline{traverse$\_$2} ( m0 ) ) )}'' for the input ``what river flows through m0''. We found that the trigram ``\texttt{\small river , 
\underline{loc$\_$2}}'' occurs 51 times in the training data while ``\texttt{\small river , 
\underline{traverse$\_$2}}'' occurs only 4 times. This is a common pattern: for 72\% of the errors from the fine-tuned T5-11B model on this split, the predicted trigram occurs more frequently  in the training data than the correct trigram.

\begin{figure}[t]
\centering
\includegraphics[width=\linewidth]{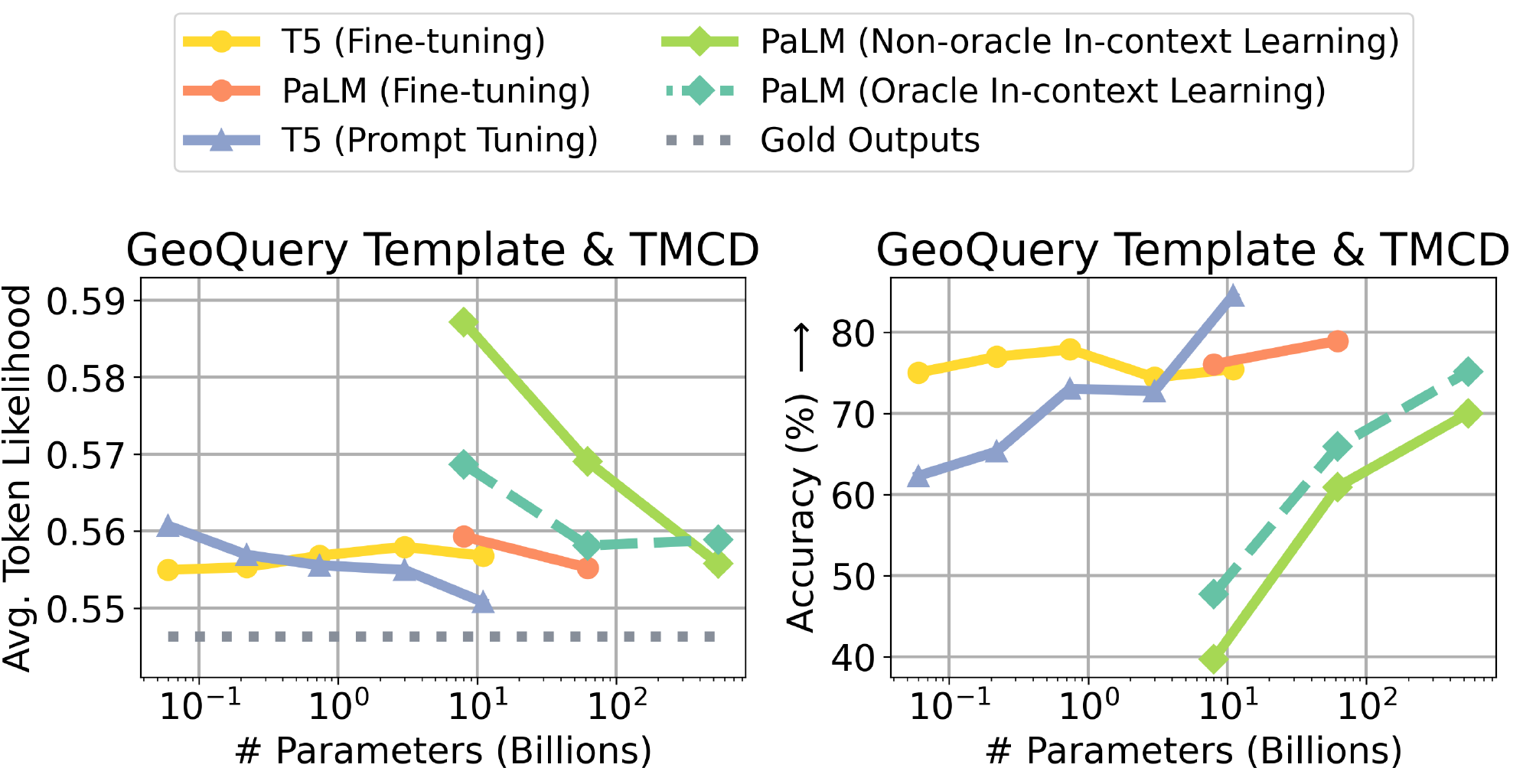}
\caption{Average token likelihood of trigram LM (left) and scaling curves (right) on development set of GeoQuery template and TMCD splits.}
\label{fig:overfit}
\end{figure}

To measure this tendency, we fit a simple count-based trigram language model (with add-1 smoothing) over outputs in the training data. We then measure the average token likelihood according to this trigram LM for the predictions compared to the gold outputs. As length extrapolation errors have been explored above, we focus on the template and TMCD splits of GeoQuery. The results are shown in Figure~\ref{fig:overfit}.
We observe that when these models make mistakes, the incorrect predictions tend to be biased towards predictions that are more likely according to the trigram LM. This suggests that such models are overfitting to these shallow statistical features to some degree. For fine-tuning, larger model scales do not necessarily alleviate this tendency. However, prompt tuning shows a more positive trend with scale. Notably, some models have lower accuracy with prompt tuning but also lower agreement with the trigram LM, suggesting that a lower proportion of the errors for prompt tuning are related to this particular type of overfitting than for fine-tuning. This overfitting issue is also reduced when increasing model size for in-context learning.

\begin{figure*}[t]
\centering
\includegraphics[width=\linewidth]{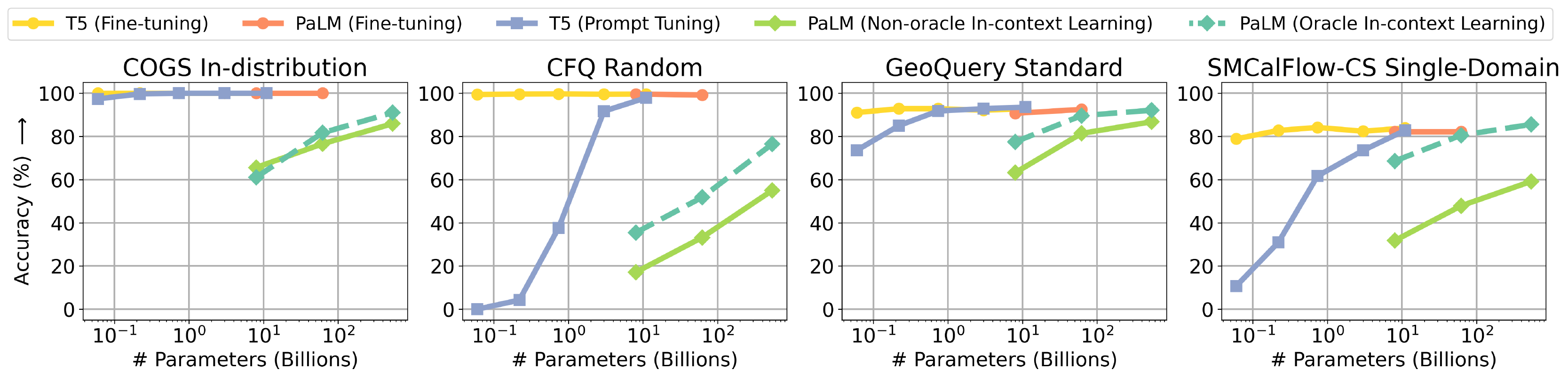}
\caption{Scaling curves for in-distribution splits of different datasets. Note that  in-context learning with an oracle retriever (dashed) cannot be compared directly with other methods as it has access to the gold output.}
\label{fig:iid_scale}
\end{figure*}

\subsection{Task Analysis}
In this section, we identify aspects of semantic parsing tasks that might contribute to scaling behaviors and analyze their impact.

\subsubsection{Distribution Shift}
We first study whether the distribution shift between training and testing affects scaling behavior. We evaluate the model performance on in-distribution splits and show results in Figure~\ref{fig:iid_scale}. We observe similar flat scaling curves when fine-tuning LMs, but the overall performance on in-distribution splits is much better than that on compositional splits. We hypothesize that fine-tuning performance on in-distribution splits does not benefit from scaling due to limited headroom (which we will investigate below). Additionally, prompt tuning is able to close the gap and matches the fine-tuning performance when model size increases. In-context learning with an oracle retriever achieves similar performance as fine-tuning and prompt tuning for non-synthetic datasets, but performs worse on synthetic datasets. In-context learning with a non-oracle retriever still lags behind both fine-tuning and prompt tuning.

\paragraph{Headroom Analysis}
We investigate whether the flat scaling curves for in-distribution evaluations are due to performance saturation as opposed to the types of error trends observed on compositional splits. For synthetic datasets, even the smallest size fine-tuned models achieve 100\% accuracy. For non-synthetic datasets, we manually sample up to 20 test examples where all models output incorrect predictions. Similar to what was observed in~\citet{qiu2022improving}, we estimate that 30\% of the errors on the GeoQuery in-distribution split and 70\% of the errors on the SMCalFlow-CS single-domain split are related to ambiguous and inconsistent annotations or unseen output symbols, suggesting limited headroom of scaling for improving performance on in-distribution splits. However, for compositional evaluations, a large number of errors are related to the error types discussed in~\S\ref{sec:error_analysis} and only around 15\% of the errors on the GeoQuery compositional split and 10\% of the errors on the SMCalFlow-CS cross-domain split are due to the dataset issues mentioned above.

As another indication of headroom, prior work shows that
extra techniques such as data augmentation could significantly boost the performance of fine-tuned LMs on compositional splits~\cite{Oren2021FindingNI, qiu2022improving}, meaning the performance of our models on compositional splits is not yet saturated.

\begin{figure}[t]
\centering
\includegraphics[width=\linewidth]{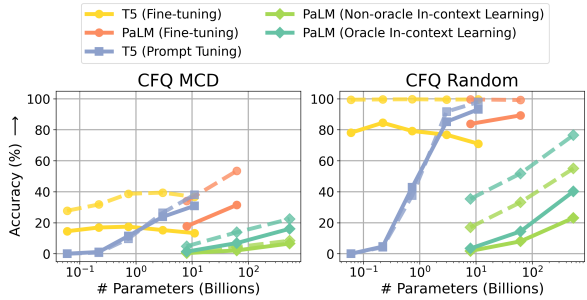}
\caption{Results on CFQ using different amount of training data. Dashed lines use the original split with around 90K training examples. Solid lines use the down-sampled split with around 1K training examples.}
\label{fig:acc_vs_data}
\end{figure}

To study the scaling behaviors on splits that are less saturated, we create smaller CFQ splits by randomly sampling 1000 examples from the original training sets. The resulting training splits can cover the required symbols in test examples with only 1--2 exceptions. From the results in Figure~\ref{fig:acc_vs_data}, the performance for all models and task adaption techniques drops when reducing the number of training examples, similar to the findings from~\citet{tsarkov2021cfq}. However, the magnitude of the difference varies across different splits and techniques. Fine-tuning performance drops significantly for both in-distribution and compositional splits. Notably, the T5 fine-tuning curve on unsaturated in-distribution split is still negative, which indicates that saturation might not be the only factor contributing to flat fine-tuning curves and more investigation is needed to provide a comprehensive explanation. In addition, prompt tuning is less sensitive to the amount of training data, demonstrating its strength when the number of training examples is limited.

\subsubsection{Output Space}
\label{sec:output_space}
\begin{figure}[t]
\centering
\includegraphics[width=\linewidth]{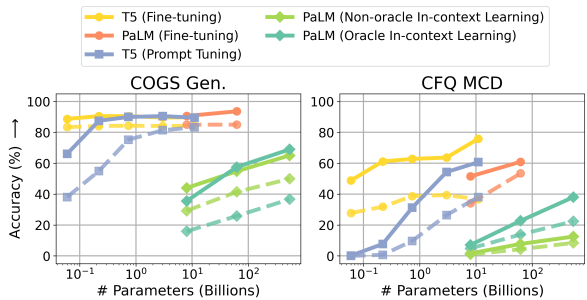}
\caption{We compare the performance using original output formats (dashed) and intermediate representations (solid). The original formats perform worse but the scaling trends are similar.}
\label{fig:output}
\end{figure}

Large LMs have shown impressive performance on generating natural language, but our study requires them to generate task-specific meaning representations that are unlikely to exist in the pre-training data. Prior work has shown the potential of leveraging alternative output representations to improve semantic parsers~\cite{herzig2021unlocking, shin-etal-2021-constrained}. We evaluate the impact of the output space using different output formats for the COGS and CFQ dataset. For COGS, we compare the original format that contains logical variables and its equivalent variable-free form. For CFQ, we compare the original format and the reversible intermediate representation from~\citet{herzig2021unlocking}.
Details of these intermediate representation can be found in Appendix~\ref{sec:appendix_ir}.
We show results\footnote{For in-context learning we use the same exemplars for both the original format and the intermediate representation for fair comparison. However, the exact number of exemplars might be different as the original format is much longer.} in Figure~\ref{fig:output}. We observe similar scaling trends for the two formats, and using the more complex original format hurts performance in all cases.

\paragraph{Error Analysis}
For COGS, when fine-tuning and prompt tuning, models struggle most on deeper recursion such as nested prepositional phrases. 
For in-context learning, in addition to deep recursion, a large number of errors are related to subtle nuances of the meaning representation. For example, the output has three types of entities (``a boy'', ``the boy'', ``James'') with different semantic forms.
While fine-tuned models successfully distinguish these, in-context learning struggles even when the exemplars include many examples of each entity type, leading to 29\% absolute accuracy loss.
The accuracy is even lower when using the original output format, which requires associating each entity with its token index. 

For CFQ, more than 60\% of the errors of fine-tuning and prompt tuning are related to missing conjuncts.
This issue is largely mitigated by using the intermediate representation that groups conjuncts and reduces the mismatch between utterances and programs~\cite{herzig2021unlocking}. However, for in-context learning, the model generates many incorrect predictions that contain conjuncts not in the ground-truth conjuncts. This reiterates the challenge of in-context learning when the output space is complex and not well represented in the pre-training data~\cite{min2022rethinking, reynolds2021prompt}.

\subsection{Retriever Analysis} 
\label{sec:retriever_analysis}

We observe that model performance when using in-context learning is strongly dependent on the method used to retrieve relevant exemplars, similar to findings from previous studies~\cite{rubin-etal-2022-learning, lu2021fantastically}.
Here we aim to better understand the differences in performance between different retrievers, which can inform future work towards improving in-context learning performance.

\paragraph{Prediction Accuracy}
We evaluate end-to-end model performance with the different retrievers and show results in Table~\ref{tab:retriever}. We find that BM25 outperforms BERT-based retriever similarly to prior work~\cite{rubin-etal-2022-learning} and that the target overlap oracle generally outperforms the BM25 oracle. The difference between using a non-oracle retriever compared to using an oracle retriever is less significant for datasets that have a smaller set of examples to select from, such as GeoQuery, than for datasets with a larger number of training examples such as SMCalFlow-CS. 

\begin{table*}[!t]
\begin{center}
\scalebox{0.82}{
\begin{tabular}{lccccccccccc}
\toprule
 & \multicolumn{2}{c}{{\bf{\textsc{COGS}}}} 
 & \multicolumn{2}{c}{{\bf{\textsc{CFQ}}}} 
 & \multicolumn{4}{c}{{\bf{\textsc{GeoQuery}}}} 
 & \multicolumn{3}{c}{{\bf\textsc{SMCalFlow-CS}}} \\
 \cmidrule(lr){2-3} \cmidrule(lr){4-5} \cmidrule(lr){6-9} \cmidrule(lr){10-12}
 & In-dist. & Gen. & Random & MCD & Std. & Templ. & TMCD & Len. & Single & Cross & Len. \\
\midrule
BERT & 85.9 & 65.0 & 57.5 & 7.8 & 86.4 & 69.7 & 63.9 & 49.1 & 56.8 & 1.4 & 9.4 \\
BM25 & 77.1 & 58.8 & 54.8 & 8.0 & 86.8 & 73.6 & 66.4 & 52.7 & 60.1 & 9.5 & 16.1 \\
\midrule
BM25 Oracle & 77.0 & 58.0 & 58.0 & 14.8 & 90.7 & 72.4 & 67.9 & 55.5 & 73.4 & 30.0 & 34,7 \\
Target Overlap Oracle & 91.0 & 69.0 & 74.6 & 21.7 & 92.1 & 74.8 & 75.5 & 56.4 & 85.6 & 43.5 & 46.4 \\
\bottomrule
\end{tabular}
}
\caption{We compare the in-context learning accuracy on development set using PaLM-540B model with different types of retrievers. We consider both non-oracle retrievers (top) and oracle retrievers (bottom). }
\label{tab:retriever}
\end{center}
\end{table*}

\begin{figure}[t]
\centering
\includegraphics[width=\linewidth]{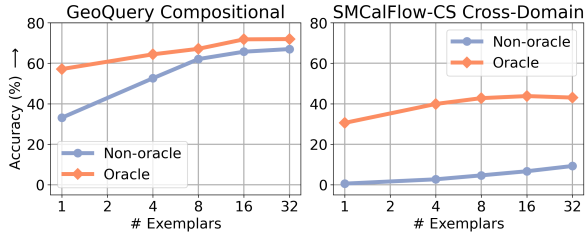}
\caption{Comparing model accuracy based on the number of exemplars with the best non-oracle retriever and oracle retriever on the development set.}
\label{fig:acc_vs_exemplars}
\end{figure}

\paragraph{Number of Exemplars}
We also study how accuracy is impacted by the number of exemplars included in prompts. We use the 540B PaLM model and show results in Figure~\ref{fig:acc_vs_exemplars}.
For both the oracle and non-oracle retrievers, performance can be improved by adding more exemplars up to a certain number; afterwards, we see that performance no longer improves on most splits. This number is greater for the non-oracle retriever than for the oracle retriever, suggesting that a smaller number of exemplars may be sufficient if the retriever is able to select the most informative examples from the training set.

\paragraph{Retriever Discussion}
For compositional generalization, retrieving the closest exemplars based on standard notions of sentence-level similarity might not be optimal.
For instance, for the query ``Schedule a meeting with my manager'' in the SMCalFlow-CS cross-domain evaluation, the highest similarity training example ``Schedule a meeting with Alex'' might be less useful than the less similar example ``Who is my manager'', as the former example lacks important cross-domain knowledge about querying an org chart. Therefore, considering alternative ways to construct prompts that balance both atom coverage and structural diversity can be an important future direction to improve compositional generalization in semantic parsing.
We also noticed high variance among permutations of a given set of exemplars, as also noted in \citet{lu2021fantastically, zhao2021calibrating}, suggesting the importance of improving the robustness and reducing the order sensitivity of prompts. We believe tailored retrieval methods and prompt design would be fruitful avenues for future research.

\section{Conclusion} 
In this paper, we study the impact of model scale on compositional generalization in semantic parsing. We evaluate encoder-decoder models up to 11B parameters and decoder-only models up to 540B parameters. We select fine-tuning, prompt tuning, and in-context learning as representative task adaptation methods for pre-trained models, covering the range of updating all to none of the model parameters for the end task.

We find that full fine-tuning generally has flat or negative scaling curves for compositional generalization evaluations, and the flat curves are not likely due to performance saturation. In-context learning can achieve performance gains with scaling, but the performance is usually worse than that of smaller fine-tuned models and highly dependent on the retriever. Prompt tuning exhibits more positive scaling curves for compositional generalization in semantic parsing and can sometimes outperform fine-tuning. 

We further conduct error analysis and identify divergent error trends. 
We find that larger models are better at modeling the output syntax, but can also suffer from overfitting, especially when fine-tuned. Using parameter-efficient task adaptation techniques such as prompt tuning could potentially improve compositional generalization with scale.
Our experiments also suggest the possibility of better leveraging scale to improve compositional generalization by designing better retrievers for in-context learning, using alternative output formats, and implementing constrained decoding to prevent invalid outputs.

\section{Limitations}
\paragraph{Models} 
Since we are studying scaling curves, we need access to models that were pre-trained in the same way but have different sizes and can be adapted to end tasks in multiple ways.
This rules out models such as GPT-3 \cite{brown2020language} which does not have public model weights. %
We did not use OPT models~\cite{zhang2022opt} as they were not available until recently.

\paragraph{Tasks}  %
Prior works have evaluated on text-to-SQL semantic parsing tasks using LMs~\cite{rajkumar2022evaluating, cheng2022binding}.
We did not evaluate on this task as it has a different task definition than the other tasks we studied. SQL generation requires conditioning on and reasoning over a database schema given as input in order to reach competitive performance. It also involves the challenge of schema matching, which could complicate the results. 

\paragraph{Runs} We did not perform multiple runs for each split due to computation costs. We aggregate results by dataset to reduce variance.
\paragraph{Task Adaptation Techniques} For the family of methods that partially update the parameters, we choose prompt tuning as a representative. We leave exploring other parameter-efficient transfer learning methods for future work.
\paragraph{Retrievers} Our results demonstrate that a better retriever maintains the scaling curve trends but improves the absolute metrics. However, designing a good retriever is outside the scope of this paper.  Our oracle retrievers are also not a true upper bound, which would require an intractable search over all possible sets of exemplars for each test query. 
\paragraph{Prompt Construction} For in-context learning, we use a fixed prompt format described in Appendix~\ref{sec:appendix_experimental_setup}, chosen based on recommended best practices and preliminary experiments. We leave exploring other prompt constructions such as providing instructions and explanations in the prompts for future work.
\paragraph{Computation Cost} Benefits of scaling up models are offset by computation costs and environmental impacts. The scaling curves with computation cost taken into account (e.g., accuracy per unit of electricity) would be an interesting direction to explore.

\section*{Acknowledgements}

We thank Ming-Wei Chang, Luheng He, Alexandre Passos, Pengcheng Yin, Yury Zemlyanskiy, the Google Research Language team, and the anonymous reviewers for helpful comments and discussions.

\bibliography{anthology,custom}

\begin{thebibliography}{83}
\expandafter\ifx\csname natexlab\endcsname\relax\def\natexlab#1{#1}\fi

\bibitem[{Abnar et~al.(2021)Abnar, Dehghani, Neyshabur, and
  Sedghi}]{abnar2021exploring}
Samira Abnar, Mostafa Dehghani, Behnam Neyshabur, and Hanie Sedghi. 2021.
\newblock \href {https://arxiv.org/abs/2110.02095} {Exploring the limits of
  large scale pre-training}.
\newblock \emph{ArXiv preprint}, abs/2110.02095.

\bibitem[{Aky{\"{u}}rek et~al.(2021)Aky{\"{u}}rek, Aky{\"{u}}rek, and
  Andreas}]{Akyrek2021LearningTR}
Ekin Aky{\"{u}}rek, Afra~Feyza Aky{\"{u}}rek, and Jacob Andreas. 2021.
\newblock \href {https://openreview.net/forum?id=PS3IMnScugk} {Learning to
  recombine and resample data for compositional generalization}.
\newblock In \emph{9th International Conference on Learning Representations,
  {ICLR} 2021, Virtual Event, Austria, May 3-7, 2021}. OpenReview.net.

\bibitem[{Andreas(2020)}]{andreas-2020-good}
Jacob Andreas. 2020.
\newblock \href {https://doi.org/10.18653/v1/2020.acl-main.676} {Good-enough
  compositional data augmentation}.
\newblock In \emph{Proceedings of the 58th Annual Meeting of the Association
  for Computational Linguistics}, pages 7556--7566, Online. Association for
  Computational Linguistics.

\bibitem[{Andreas et~al.(2020)Andreas, Bufe, Burkett, Chen, Clausman, Crawford,
  Crim, DeLoach, Dorner, Eisner, Fang, Guo, Hall, Hayes, Hill, Ho, Iwaszuk,
  Jha, Klein, Krishnamurthy, Lanman, Liang, Lin, Lintsbakh, McGovern,
  Nisnevich, Pauls, Petters, Read, Roth, Roy, Rusak, Short, Slomin, Snyder,
  Striplin, Su, Tellman, Thomson, Vorobev, Witoszko, Wolfe, Wray, Zhang, and
  Zotov}]{andreas-etal-2020-task}
Jacob Andreas, John Bufe, David Burkett, Charles Chen, Josh Clausman, Jean
  Crawford, Kate Crim, Jordan DeLoach, Leah Dorner, Jason Eisner, Hao Fang,
  Alan Guo, David Hall, Kristin Hayes, Kellie Hill, Diana Ho, Wendy Iwaszuk,
  Smriti Jha, Dan Klein, Jayant Krishnamurthy, Theo Lanman, Percy Liang,
  Christopher~H. Lin, Ilya Lintsbakh, Andy McGovern, Aleksandr Nisnevich, Adam
  Pauls, Dmitrij Petters, Brent Read, Dan Roth, Subhro Roy, Jesse Rusak, Beth
  Short, Div Slomin, Ben Snyder, Stephon Striplin, Yu~Su, Zachary Tellman, Sam
  Thomson, Andrei Vorobev, Izabela Witoszko, Jason Wolfe, Abby Wray, Yuchen
  Zhang, and Alexander Zotov. 2020.
\newblock \href {https://doi.org/10.1162/tacl_a_00333} {Task-oriented dialogue
  as dataflow synthesis}.
\newblock \emph{Transactions of the Association for Computational Linguistics},
  8:556--571.

\bibitem[{Austin et~al.(2021)Austin, Odena, Nye, Bosma, Michalewski, Dohan,
  Jiang, Cai, Terry, Le, and Sutton}]{austin2021program}
Jacob Austin, Augustus Odena, Maxwell~I. Nye, Maarten Bosma, Henryk
  Michalewski, David Dohan, Ellen Jiang, Carrie~J. Cai, Michael Terry, Quoc~V.
  Le, and Charles Sutton. 2021.
\newblock \href {https://arxiv.org/abs/2108.07732} {Program synthesis with
  large language models}.
\newblock \emph{ArXiv preprint}, abs/2108.07732.

\bibitem[{Bahri et~al.(2021)Bahri, Dyer, Kaplan, Lee, and
  Sharma}]{bahri2021explaining}
Yasaman Bahri, Ethan Dyer, Jared Kaplan, Jaehoon Lee, and Utkarsh Sharma. 2021.
\newblock \href {https://arxiv.org/abs/2102.06701} {Explaining neural scaling
  laws}.
\newblock \emph{ArXiv preprint}, abs/2102.06701.

\bibitem[{Battaglia et~al.(2018)Battaglia, Hamrick, Bapst, Sanchez-Gonzalez,
  Zambaldi, Malinowski, Tacchetti, Raposo, Santoro, Faulkner
  et~al.}]{battaglia2018relational}
Peter~W Battaglia, Jessica~B Hamrick, Victor Bapst, Alvaro Sanchez-Gonzalez,
  Vinicius Zambaldi, Mateusz Malinowski, Andrea Tacchetti, David Raposo, Adam
  Santoro, Ryan Faulkner, et~al. 2018.
\newblock \href {https://arxiv.org/abs/1806.01261} {Relational inductive
  biases, deep learning, and graph networks}.
\newblock \emph{ArXiv preprint}, abs/1806.01261.

\bibitem[{Bornschein et~al.(2020)Bornschein, Visin, and
  Osindero}]{bornschein2020small}
J{\"{o}}rg Bornschein, Francesco Visin, and Simon Osindero. 2020.
\newblock \href {http://proceedings.mlr.press/v119/bornschein20a.html} {Small
  data, big decisions: Model selection in the small-data regime}.
\newblock In \emph{Proceedings of the 37th International Conference on Machine
  Learning, {ICML} 2020, 13-18 July 2020, Virtual Event}, volume 119 of
  \emph{Proceedings of Machine Learning Research}, pages 1035--1044. {PMLR}.

\bibitem[{Brown et~al.(2020)Brown, Mann, Ryder, Subbiah, Kaplan, Dhariwal,
  Neelakantan, Shyam, Sastry, Askell, Agarwal, Herbert{-}Voss, Krueger,
  Henighan, Child, Ramesh, Ziegler, Wu, Winter, Hesse, Chen, Sigler, Litwin,
  Gray, Chess, Clark, Berner, McCandlish, Radford, Sutskever, and
  Amodei}]{brown2020language}
Tom~B. Brown, Benjamin Mann, Nick Ryder, Melanie Subbiah, Jared Kaplan,
  Prafulla Dhariwal, Arvind Neelakantan, Pranav Shyam, Girish Sastry, Amanda
  Askell, Sandhini Agarwal, Ariel Herbert{-}Voss, Gretchen Krueger, Tom
  Henighan, Rewon Child, Aditya Ramesh, Daniel~M. Ziegler, Jeffrey Wu, Clemens
  Winter, Christopher Hesse, Mark Chen, Eric Sigler, Mateusz Litwin, Scott
  Gray, Benjamin Chess, Jack Clark, Christopher Berner, Sam McCandlish, Alec
  Radford, Ilya Sutskever, and Dario Amodei. 2020.
\newblock \href
  {https://proceedings.neurips.cc/paper/2020/hash/1457c0d6bfcb4967418bfb8ac142f64a-Abstract.html}
  {Language models are few-shot learners}.
\newblock In \emph{Advances in Neural Information Processing Systems 33: Annual
  Conference on Neural Information Processing Systems 2020, NeurIPS 2020,
  December 6-12, 2020, virtual}.

\bibitem[{Chen et~al.(2020)Chen, Liang, Yu, Song, and
  Zhou}]{chen2020compositional}
Xinyun Chen, Chen Liang, Adams~Wei Yu, Dawn Song, and Denny Zhou. 2020.
\newblock \href
  {https://proceedings.neurips.cc/paper/2020/hash/12b1e42dc0746f22cf361267de07073f-Abstract.html}
  {Compositional generalization via neural-symbolic stack machines}.
\newblock In \emph{Advances in Neural Information Processing Systems 33: Annual
  Conference on Neural Information Processing Systems 2020, NeurIPS 2020,
  December 6-12, 2020, virtual}.

\bibitem[{Cheng et~al.(2022)Cheng, Xie, Shi, Li, Nadkarni, Hu, Xiong, Radev,
  Ostendorf, Zettlemoyer, Smith, and Yu}]{cheng2022binding}
Zhoujun Cheng, Tianbao Xie, Peng Shi, Chengzu Li, Rahul Nadkarni, Yushi Hu,
  Caiming Xiong, Dragomir Radev, Mari Ostendorf, Luke Zettlemoyer, Noah~A.
  Smith, and Tao Yu. 2022.
\newblock \href {https://arxiv.org/abs/2210.02875} {Binding language models in
  symbolic languages}.
\newblock \emph{ArXiv preprint}, abs/2210.02875.

\bibitem[{Chowdhery et~al.(2022)Chowdhery, Narang, Devlin, Bosma, Mishra,
  Roberts, Barham, Chung, Sutton, Gehrmann, Schuh, Shi, Tsvyashchenko, Maynez,
  Rao, Barnes, Tay, Shazeer, Prabhakaran, Reif, Du, Hutchinson, Pope, Bradbury,
  Austin, Isard, Gur-Ari, Yin, Duke, Levskaya, Ghemawat, Dev, Michalewski,
  Garcia, Misra, Robinson, Fedus, Zhou, Ippolito, Luan, Lim, Zoph, Spiridonov,
  Sepassi, Dohan, Agrawal, Omernick, Dai, Pillai, Pellat, Lewkowycz, Moreira,
  Child, Polozov, Lee, Zhou, Wang, Saeta, Diaz, Firat, Catasta, Wei,
  Meier-Hellstern, Eck, Dean, Petrov, and Fiedel}]{chowdhery2022palm}
Aakanksha Chowdhery, Sharan Narang, Jacob Devlin, Maarten Bosma, Gaurav Mishra,
  Adam Roberts, Paul Barham, Hyung~Won Chung, Charles Sutton, Sebastian
  Gehrmann, Parker Schuh, Kensen Shi, Sasha Tsvyashchenko, Joshua Maynez,
  Abhishek Rao, Parker Barnes, Yi~Tay, Noam Shazeer, Vinodkumar Prabhakaran,
  Emily Reif, Nan Du, Ben Hutchinson, Reiner Pope, James Bradbury, Jacob
  Austin, Michael Isard, Guy Gur-Ari, Pengcheng Yin, Toju Duke, Anselm
  Levskaya, Sanjay Ghemawat, Sunipa Dev, Henryk Michalewski, Xavier Garcia,
  Vedant Misra, Kevin Robinson, Liam Fedus, Denny Zhou, Daphne Ippolito, David
  Luan, Hyeontaek Lim, Barret Zoph, Alexander Spiridonov, Ryan Sepassi, David
  Dohan, Shivani Agrawal, Mark Omernick, Andrew~M. Dai,
  Thanumalayan~Sankaranarayana Pillai, Marie Pellat, Aitor Lewkowycz, Erica
  Moreira, Rewon Child, Oleksandr Polozov, Katherine Lee, Zongwei Zhou, Xuezhi
  Wang, Brennan Saeta, Mark Diaz, Orhan Firat, Michele Catasta, Jason Wei,
  Kathy Meier-Hellstern, Douglas Eck, Jeff Dean, Slav Petrov, and Noah Fiedel.
  2022.
\newblock \href {https://arxiv.org/abs/2204.02311} {Palm: Scaling language
  modeling with pathways}.
\newblock \emph{ArXiv preprint}, abs/2204.02311.

\bibitem[{Conklin et~al.(2021)Conklin, Wang, Smith, and
  Titov}]{conklin2021meta}
Henry Conklin, Bailin Wang, Kenny Smith, and Ivan Titov. 2021.
\newblock \href {https://doi.org/10.18653/v1/2021.acl-long.258} {Meta-learning
  to compositionally generalize}.
\newblock In \emph{Proceedings of the 59th Annual Meeting of the Association
  for Computational Linguistics and the 11th International Joint Conference on
  Natural Language Processing (Volume 1: Long Papers)}, pages 3322--3335,
  Online. Association for Computational Linguistics.

\bibitem[{Csord{\'a}s et~al.(2021)Csord{\'a}s, Irie, and
  Schmidhuber}]{csordas2021devil}
R{\'o}bert Csord{\'a}s, Kazuki Irie, and Juergen Schmidhuber. 2021.
\newblock \href {https://doi.org/10.18653/v1/2021.emnlp-main.49} {The devil is
  in the detail: Simple tricks improve systematic generalization of
  transformers}.
\newblock In \emph{Proceedings of the 2021 Conference on Empirical Methods in
  Natural Language Processing}, pages 619--634, Online and Punta Cana,
  Dominican Republic. Association for Computational Linguistics.

\bibitem[{Devlin et~al.(2019)Devlin, Chang, Lee, and
  Toutanova}]{devlin2018bert}
Jacob Devlin, Ming-Wei Chang, Kenton Lee, and Kristina Toutanova. 2019.
\newblock \href {https://doi.org/10.18653/v1/N19-1423} {{BERT}: Pre-training of
  deep bidirectional transformers for language understanding}.
\newblock In \emph{Proceedings of the 2019 Conference of the North {A}merican
  Chapter of the Association for Computational Linguistics: Human Language
  Technologies, Volume 1 (Long and Short Papers)}, pages 4171--4186,
  Minneapolis, Minnesota. Association for Computational Linguistics.

\bibitem[{Finegan-Dollak et~al.(2018)Finegan-Dollak, Kummerfeld, Zhang,
  Ramanathan, Sadasivam, Zhang, and Radev}]{finegan2018improving}
Catherine Finegan-Dollak, Jonathan~K. Kummerfeld, Li~Zhang, Karthik Ramanathan,
  Sesh Sadasivam, Rui Zhang, and Dragomir Radev. 2018.
\newblock \href {https://doi.org/10.18653/v1/P18-1033} {Improving text-to-{SQL}
  evaluation methodology}.
\newblock In \emph{Proceedings of the 56th Annual Meeting of the Association
  for Computational Linguistics (Volume 1: Long Papers)}, pages 351--360,
  Melbourne, Australia. Association for Computational Linguistics.

\bibitem[{Furrer et~al.(2020)Furrer, van Zee, Scales, and
  Sch{\"a}rli}]{furrer2020compositional}
Daniel Furrer, Marc van Zee, Nathan Scales, and Nathanael Sch{\"a}rli. 2020.
\newblock \href {https://arxiv.org/abs/2007.08970} {Compositional
  generalization in semantic parsing: Pre-training vs. specialized
  architectures}.
\newblock \emph{ArXiv preprint}, abs/2007.08970.

\bibitem[{Geiger et~al.(2019)Geiger, Jacot, Spigler, Gabriel, Sagun, d'Ascoli,
  Biroli, Hongler, and Wyart}]{geiger2019scaling}
Mario Geiger, Arthur Jacot, Stefano Spigler, Franck Gabriel, Levent Sagun,
  St{\'{e}}phane d'Ascoli, Giulio Biroli, Cl{\'{e}}ment Hongler, and Matthieu
  Wyart. 2019.
\newblock \href {https://arxiv.org/abs/1901.01608} {Scaling description of
  generalization with number of parameters in deep learning}.
\newblock \emph{ArXiv preprint}, abs/1901.01608.

\bibitem[{Ghorbani et~al.(2021)Ghorbani, Firat, Freitag, Bapna, Krikun, Garcia,
  Chelba, and Cherry}]{ghorbani2021scaling}
Behrooz Ghorbani, Orhan Firat, Markus Freitag, Ankur Bapna, Maxim Krikun,
  Xavier Garcia, Ciprian Chelba, and Colin Cherry. 2021.
\newblock \href {https://arxiv.org/abs/2109.07740} {Scaling laws for neural
  machine translation}.
\newblock \emph{ArXiv preprint}, abs/2109.07740.

\bibitem[{Gordon et~al.(2020)Gordon, Lopez{-}Paz, Baroni, and
  Bouchacourt}]{gordon2020permutation}
Jonathan Gordon, David Lopez{-}Paz, Marco Baroni, and Diane Bouchacourt. 2020.
\newblock \href {https://openreview.net/forum?id=SylVNerFvr} {Permutation
  equivariant models for compositional generalization in language}.
\newblock In \emph{8th International Conference on Learning Representations,
  {ICLR} 2020, Addis Ababa, Ethiopia, April 26-30, 2020}. OpenReview.net.

\bibitem[{He et~al.(2022)He, Zhou, Ma, Berg-Kirkpatrick, and
  Neubig}]{he2022towards}
Junxian He, Chunting Zhou, Xuezhe Ma, Taylor Berg-Kirkpatrick, and Graham
  Neubig. 2022.
\newblock \href {https://openreview.net/forum?id=0RDcd5Axok} {Towards a unified
  view of parameter-efficient transfer learning}.
\newblock In \emph{International Conference on Learning Representations}.

\bibitem[{Henighan et~al.(2020)Henighan, Kaplan, Katz, Chen, Hesse, Jackson,
  Jun, Brown, Dhariwal, Gray, Hallacy, Mann, Radford, Ramesh, Ryder, Ziegler,
  Schulman, Amodei, and McCandlish}]{henighan2020scaling}
Tom Henighan, Jared Kaplan, Mor Katz, Mark Chen, Christopher Hesse, Jacob
  Jackson, Heewoo Jun, Tom~B. Brown, Prafulla Dhariwal, Scott Gray, Chris
  Hallacy, Benjamin Mann, Alec Radford, Aditya Ramesh, Nick Ryder, Daniel~M.
  Ziegler, John Schulman, Dario Amodei, and Sam McCandlish. 2020.
\newblock \href {https://arxiv.org/abs/2010.14701} {Scaling laws for
  autoregressive generative modeling}.
\newblock \emph{ArXiv preprint}, abs/2010.14701.

\bibitem[{Hernandez et~al.(2021)Hernandez, Kaplan, Henighan, and
  McCandlish}]{hern2021scaling}
Danny Hernandez, Jared Kaplan, Tom Henighan, and Sam McCandlish. 2021.
\newblock \href {http://arxiv.org/abs/2102.01293} {Scaling laws for transfer}.
\newblock \emph{CoRR}.

\bibitem[{Herzig and Berant(2019)}]{herzig2019don}
Jonathan Herzig and Jonathan Berant. 2019.
\newblock \href {https://doi.org/10.18653/v1/D19-1394} {Don{'}t paraphrase,
  detect! rapid and effective data collection for semantic parsing}.
\newblock In \emph{Proceedings of the 2019 Conference on Empirical Methods in
  Natural Language Processing and the 9th International Joint Conference on
  Natural Language Processing (EMNLP-IJCNLP)}, pages 3810--3820, Hong Kong,
  China. Association for Computational Linguistics.

\bibitem[{Herzig and Berant(2021)}]{herzig2020span}
Jonathan Herzig and Jonathan Berant. 2021.
\newblock \href {https://doi.org/10.18653/v1/2021.acl-long.74} {Span-based
  semantic parsing for compositional generalization}.
\newblock In \emph{Proceedings of the 59th Annual Meeting of the Association
  for Computational Linguistics and the 11th International Joint Conference on
  Natural Language Processing (Volume 1: Long Papers)}, pages 908--921, Online.
  Association for Computational Linguistics.

\bibitem[{Herzig et~al.(2021)Herzig, Shaw, Chang, Guu, Pasupat, and
  Zhang}]{herzig2021unlocking}
Jonathan Herzig, Peter Shaw, Ming-Wei Chang, Kelvin Guu, Panupong Pasupat, and
  Yuan Zhang. 2021.
\newblock \href {https://arxiv.org/abs/2104.07478} {Unlocking compositional
  generalization in pre-trained models using intermediate representations}.
\newblock \emph{ArXiv preprint}, abs/2104.07478.

\bibitem[{Hestness et~al.(2017)Hestness, Narang, Ardalani, Diamos, Jun,
  Kianinejad, Patwary, Yang, and Zhou}]{hestness2017deep}
Joel Hestness, Sharan Narang, Newsha Ardalani, Gregory~F. Diamos, Heewoo Jun,
  Hassan Kianinejad, Md. Mostofa~Ali Patwary, Yang Yang, and Yanqi Zhou. 2017.
\newblock \href {https://arxiv.org/abs/1712.00409} {Deep learning scaling is
  predictable, empirically}.
\newblock \emph{ArXiv preprint}, abs/1712.00409.

\bibitem[{Hoffmann et~al.(2022)Hoffmann, Borgeaud, Mensch, Buchatskaya, Cai,
  Rutherford, de~Las~Casas, Hendricks, Welbl, Clark, Hennigan, Noland,
  Millican, van~den Driessche, Damoc, Guy, Osindero, Simonyan, Elsen, Rae,
  Vinyals, and Sifre}]{hoffmann2022training}
Jordan Hoffmann, Sebastian Borgeaud, Arthur Mensch, Elena Buchatskaya, Trevor
  Cai, Eliza Rutherford, Diego de~Las~Casas, Lisa~Anne Hendricks, Johannes
  Welbl, Aidan Clark, Tom Hennigan, Eric Noland, Katie Millican, George van~den
  Driessche, Bogdan Damoc, Aurelia Guy, Simon Osindero, Karen Simonyan, Erich
  Elsen, Jack~W. Rae, Oriol Vinyals, and Laurent Sifre. 2022.
\newblock \href {https://doi.org/10.48550/arXiv.2203.15556} {Training
  compute-optimal large language models}.
\newblock \emph{CoRR}, abs/2203.15556.

\bibitem[{Ivgi et~al.(2022)Ivgi, Carmon, and Berant}]{ivgi2022scaling}
Maor Ivgi, Yair Carmon, and Jonathan Berant. 2022.
\newblock Scaling laws under the microscope: Predicting transformer performance
  from small scale experiments.
\newblock \emph{arXiv preprint arXiv:2202.06387}.

\bibitem[{Jia and Liang(2016)}]{jia2016data}
Robin Jia and Percy Liang. 2016.
\newblock \href {https://doi.org/10.18653/v1/P16-1002} {Data recombination for
  neural semantic parsing}.
\newblock In \emph{Proceedings of the 54th Annual Meeting of the Association
  for Computational Linguistics (Volume 1: Long Papers)}, pages 12--22, Berlin,
  Germany. Association for Computational Linguistics.

\bibitem[{Jiang and Bansal(2021)}]{jiang2021inducing}
Yichen Jiang and Mohit Bansal. 2021.
\newblock \href {https://doi.org/10.18653/v1/2021.emnlp-main.505} {Inducing
  transformer{'}s compositional generalization ability via auxiliary sequence
  prediction tasks}.
\newblock In \emph{Proceedings of the 2021 Conference on Empirical Methods in
  Natural Language Processing}, pages 6253--6265, Online and Punta Cana,
  Dominican Republic. Association for Computational Linguistics.

\bibitem[{Kaplan et~al.(2020)Kaplan, McCandlish, Henighan, Brown, Chess, Child,
  andkaplan2020scaling Alec~Radford, Wu, and Amodei}]{kaplan2020scaling}
Jared Kaplan, Sam McCandlish, Tom Henighan, Tom~B. Brown, Benjamin Chess, Rewon
  Child, Scott~Gray andkaplan2020scaling Alec~Radford, Jeffrey Wu, and Dario
  Amodei. 2020.
\newblock \href {https://arxiv.org/abs/2001.08361} {Scaling laws for neural
  language models}.
\newblock \emph{ArXiv preprint}, abs/2001.08361.

\bibitem[{Keysers et~al.(2020)Keysers, Sch{\"{a}}rli, Scales, Buisman, Furrer,
  Kashubin, Momchev, Sinopalnikov, Stafiniak, Tihon, Tsarkov, Wang, van Zee,
  and Bousquet}]{keysers2019measuring}
Daniel Keysers, Nathanael Sch{\"{a}}rli, Nathan Scales, Hylke Buisman, Daniel
  Furrer, Sergii Kashubin, Nikola Momchev, Danila Sinopalnikov, Lukasz
  Stafiniak, Tibor Tihon, Dmitry Tsarkov, Xiao Wang, Marc van Zee, and Olivier
  Bousquet. 2020.
\newblock \href {https://openreview.net/forum?id=SygcCnNKwr} {Measuring
  compositional generalization: {A} comprehensive method on realistic data}.
\newblock In \emph{8th International Conference on Learning Representations,
  {ICLR} 2020, Addis Ababa, Ethiopia, April 26-30, 2020}. OpenReview.net.

\bibitem[{Kim and Linzen(2020)}]{kim-linzen-2020-cogs}
Najoung Kim and Tal Linzen. 2020.
\newblock \href {https://doi.org/10.18653/v1/2020.emnlp-main.731} {{COGS}: A
  compositional generalization challenge based on semantic interpretation}.
\newblock In \emph{Proceedings of the 2020 Conference on Empirical Methods in
  Natural Language Processing (EMNLP)}, pages 9087--9105, Online. Association
  for Computational Linguistics.

\bibitem[{Kumar et~al.(2022)Kumar, Raghunathan, Jones, Ma, and
  Liang}]{kumar2022finetuning}
Ananya Kumar, Aditi Raghunathan, Robbie Jones, Tengyu Ma, and Percy Liang.
  2022.
\newblock \href {https://arxiv.org/abs/2202.10054} {Fine-tuning can distort
  pretrained features and underperform out-of-distribution}.
\newblock \emph{ArXiv preprint}, abs/2202.10054.

\bibitem[{Lake(2019)}]{lake2019compositional}
Brenden~M. Lake. 2019.
\newblock \href
  {https://proceedings.neurips.cc/paper/2019/hash/f4d0e2e7fc057a58f7ca4a391f01940a-Abstract.html}
  {Compositional generalization through meta sequence-to-sequence learning}.
\newblock In \emph{Advances in Neural Information Processing Systems 32: Annual
  Conference on Neural Information Processing Systems 2019, NeurIPS 2019,
  December 8-14, 2019, Vancouver, BC, Canada}, pages 9788--9798.

\bibitem[{Lake and Baroni(2018)}]{lake2018generalization}
Brenden~M. Lake and Marco Baroni. 2018.
\newblock \href {http://proceedings.mlr.press/v80/lake18a.html} {Generalization
  without systematicity: On the compositional skills of sequence-to-sequence
  recurrent networks}.
\newblock In \emph{Proceedings of the 35th International Conference on Machine
  Learning, {ICML} 2018, Stockholmsm{\"{a}}ssan, Stockholm, Sweden, July 10-15,
  2018}, volume~80 of \emph{Proceedings of Machine Learning Research}, pages
  2879--2888. {PMLR}.

\bibitem[{Lake et~al.(2017)Lake, Ullman, Tenenbaum, and
  Gershman}]{lake2017building}
Brenden~M Lake, Tomer~D Ullman, Joshua~B Tenenbaum, and Samuel~J Gershman.
  2017.
\newblock Building machines that learn and think like people.
\newblock \emph{Behavioral and brain sciences}, 40.

\bibitem[{Lester et~al.(2021)Lester, Al-Rfou, and
  Constant}]{lester-etal-2021-power}
Brian Lester, Rami Al-Rfou, and Noah Constant. 2021.
\newblock \href {https://doi.org/10.18653/v1/2021.emnlp-main.243} {The power of
  scale for parameter-efficient prompt tuning}.
\newblock In \emph{Proceedings of the 2021 Conference on Empirical Methods in
  Natural Language Processing}, pages 3045--3059, Online and Punta Cana,
  Dominican Republic. Association for Computational Linguistics.

\bibitem[{Lewis et~al.(2020)Lewis, Liu, Goyal, Ghazvininejad, Mohamed, Levy,
  Stoyanov, and Zettlemoyer}]{lewis-etal-2020-bart}
Mike Lewis, Yinhan Liu, Naman Goyal, Marjan Ghazvininejad, Abdelrahman Mohamed,
  Omer Levy, Veselin Stoyanov, and Luke Zettlemoyer. 2020.
\newblock \href {https://doi.org/10.18653/v1/2020.acl-main.703} {{BART}:
  Denoising sequence-to-sequence pre-training for natural language generation,
  translation, and comprehension}.
\newblock In \emph{Proceedings of the 58th Annual Meeting of the Association
  for Computational Linguistics}, pages 7871--7880, Online. Association for
  Computational Linguistics.

\bibitem[{Li and Liang(2021)}]{li-liang-2021-prefix}
Xiang~Lisa Li and Percy Liang. 2021.
\newblock \href {https://doi.org/10.18653/v1/2021.acl-long.353} {Prefix-tuning:
  Optimizing continuous prompts for generation}.
\newblock In \emph{Proceedings of the 59th Annual Meeting of the Association
  for Computational Linguistics and the 11th International Joint Conference on
  Natural Language Processing (Volume 1: Long Papers)}, pages 4582--4597,
  Online. Association for Computational Linguistics.

\bibitem[{Li et~al.(2019)Li, Zhao, Wang, and Hestness}]{li2019compositional}
Yuanpeng Li, Liang Zhao, Jianyu Wang, and Joel Hestness. 2019.
\newblock \href {https://doi.org/10.18653/v1/D19-1438} {Compositional
  generalization for primitive substitutions}.
\newblock In \emph{Proceedings of the 2019 Conference on Empirical Methods in
  Natural Language Processing and the 9th International Joint Conference on
  Natural Language Processing (EMNLP-IJCNLP)}, pages 4293--4302, Hong Kong,
  China. Association for Computational Linguistics.

\bibitem[{Liu et~al.(2021{\natexlab{a}})Liu, Yuan, Fu, Jiang, Hayashi, and
  Neubig}]{ppp_survey_neubig}
Pengfei Liu, Weizhe Yuan, Jinlan Fu, Zhengbao Jiang, Hiroaki Hayashi, and
  Graham Neubig. 2021{\natexlab{a}}.
\newblock \href {https://arxiv.org/abs/2107.13586} {Pre-train, prompt, and
  predict: {A} systematic survey of prompting methods in natural language
  processing}.
\newblock \emph{ArXiv preprint}, abs/2107.13586.

\bibitem[{Liu et~al.(2020)Liu, An, Lou, Chen, Lin, Gao, Zhou, Zheng, and
  Zhang}]{liu2020compositional}
Qian Liu, Shengnan An, Jian{-}Guang Lou, Bei Chen, Zeqi Lin, Yan Gao, Bin Zhou,
  Nanning Zheng, and Dongmei Zhang. 2020.
\newblock \href
  {https://proceedings.neurips.cc/paper/2020/hash/83adc9225e4deb67d7ce42d58fe5157c-Abstract.html}
  {Compositional generalization by learning analytical expressions}.
\newblock In \emph{Advances in Neural Information Processing Systems 33: Annual
  Conference on Neural Information Processing Systems 2020, NeurIPS 2020,
  December 6-12, 2020, virtual}.

\bibitem[{Liu et~al.(2021{\natexlab{b}})Liu, Ji, Fu, Du, Yang, and
  Tang}]{liu2021ptuning}
Xiao Liu, Kaixuan Ji, Yicheng Fu, Zhengxiao Du, Zhilin Yang, and Jie Tang.
  2021{\natexlab{b}}.
\newblock \href {https://arxiv.org/abs/2110.07602} {P-tuning v2: Prompt tuning
  can be comparable to fine-tuning universally across scales and tasks}.
\newblock \emph{ArXiv preprint}, abs/2110.07602.

\bibitem[{Liu et~al.(2019)Liu, Ott, Goyal, Du, Joshi, Chen, Levy, Lewis,
  Zettlemoyer, and Stoyanov}]{liu2019roberta}
Yinhan Liu, Myle Ott, Naman Goyal, Jingfei Du, Mandar Joshi, Danqi Chen, Omer
  Levy, Mike Lewis, Luke Zettlemoyer, and Veselin Stoyanov. 2019.
\newblock \href {http://arxiv.org/abs/1907.11692} {Roberta: A robustly
  optimized bert pretraining approach}.
\newblock \emph{CoRR}.

\bibitem[{Lu et~al.(2021)Lu, Bartolo, Moore, Riedel, and
  Stenetorp}]{lu2021fantastically}
Yao Lu, Max Bartolo, Alastair Moore, Sebastian Riedel, and Pontus Stenetorp.
  2021.
\newblock \href {https://arxiv.org/abs/2104.08786} {Fantastically ordered
  prompts and where to find them: Overcoming few-shot prompt order
  sensitivity}.
\newblock \emph{ArXiv preprint}, abs/2104.08786.

\bibitem[{Min et~al.(2022)Min, Lyu, Holtzman, Artetxe, Lewis, Hajishirzi, and
  Zettlemoyer}]{min2022rethinking}
Sewon Min, Xinxi Lyu, Ari Holtzman, Mikel Artetxe, Mike Lewis, Hannaneh
  Hajishirzi, and Luke Zettlemoyer. 2022.
\newblock \href {https://arxiv.org/abs/2202.12837} {Rethinking the role of
  demonstrations: What makes in-context learning work?}
\newblock \emph{ArXiv preprint}, abs/2202.12837.

\bibitem[{Newman et~al.(2020)Newman, Hewitt, Liang, and
  Manning}]{newman2020eos}
Benjamin Newman, John Hewitt, Percy Liang, and Christopher~D. Manning. 2020.
\newblock \href {https://doi.org/10.18653/v1/2020.blackboxnlp-1.26} {The {EOS}
  decision and length extrapolation}.
\newblock In \emph{Proceedings of the Third BlackboxNLP Workshop on Analyzing
  and Interpreting Neural Networks for NLP}, pages 276--291, Online.
  Association for Computational Linguistics.

\bibitem[{Nye et~al.(2020)Nye, Solar{-}Lezama, Tenenbaum, and
  Lake}]{nye2020learning}
Maxwell~I. Nye, Armando Solar{-}Lezama, Josh Tenenbaum, and Brenden~M. Lake.
  2020.
\newblock \href
  {https://proceedings.neurips.cc/paper/2020/hash/7a685d9edd95508471a9d3d6fcace432-Abstract.html}
  {Learning compositional rules via neural program synthesis}.
\newblock In \emph{Advances in Neural Information Processing Systems 33: Annual
  Conference on Neural Information Processing Systems 2020, NeurIPS 2020,
  December 6-12, 2020, virtual}.

\bibitem[{Ontan{\'o}n et~al.(2021)Ontan{\'o}n, Ainslie, Cvicek, and
  Fisher}]{ontanon2021making}
Santiago Ontan{\'o}n, Joshua Ainslie, Vaclav Cvicek, and Zachary Fisher. 2021.
\newblock \href {https://arxiv.org/abs/2108.04378} {Making transformers solve
  compositional tasks}.
\newblock \emph{ArXiv preprint}, abs/2108.04378.

\bibitem[{Oren et~al.(2021)Oren, Herzig, and Berant}]{Oren2021FindingNI}
Inbar Oren, Jonathan Herzig, and Jonathan Berant. 2021.
\newblock \href {https://doi.org/10.18653/v1/2021.emnlp-main.843} {Finding
  needles in a haystack: Sampling structurally-diverse training sets from
  synthetic data for compositional generalization}.
\newblock In \emph{Proceedings of the 2021 Conference on Empirical Methods in
  Natural Language Processing}, pages 10793--10809, Online and Punta Cana,
  Dominican Republic. Association for Computational Linguistics.

\bibitem[{Oren et~al.(2020)Oren, Herzig, Gupta, Gardner, and
  Berant}]{oren2020improving}
Inbar Oren, Jonathan Herzig, Nitish Gupta, Matt Gardner, and Jonathan Berant.
  2020.
\newblock \href {https://doi.org/10.18653/v1/2020.findings-emnlp.225}
  {Improving compositional generalization in semantic parsing}.
\newblock In \emph{Findings of the Association for Computational Linguistics:
  EMNLP 2020}, pages 2482--2495, Online. Association for Computational
  Linguistics.

\bibitem[{Platanios et~al.(2021)Platanios, Pauls, Roy, Zhang, Kyte, Guo,
  Thomson, Krishnamurthy, Wolfe, Andreas, and
  Klein}]{platanios-etal-2021-value}
Emmanouil~Antonios Platanios, Adam Pauls, Subhro Roy, Yuchen Zhang, Alexander
  Kyte, Alan Guo, Sam Thomson, Jayant Krishnamurthy, Jason Wolfe, Jacob
  Andreas, and Dan Klein. 2021.
\newblock \href {https://doi.org/10.18653/v1/2021.acl-long.284} {Value-agnostic
  conversational semantic parsing}.
\newblock In \emph{Proceedings of the 59th Annual Meeting of the Association
  for Computational Linguistics and the 11th International Joint Conference on
  Natural Language Processing (Volume 1: Long Papers)}, pages 3666--3681,
  Online. Association for Computational Linguistics.

\bibitem[{Qiu et~al.(2022)Qiu, Shaw, Pasupat, Nowak, Linzen, Sha, and
  Toutanova}]{qiu2022improving}
Linlu Qiu, Peter Shaw, Panupong Pasupat, Pawel Nowak, Tal Linzen, Fei Sha, and
  Kristina Toutanova. 2022.
\newblock \href {https://doi.org/10.18653/v1/2022.naacl-main.323} {Improving
  compositional generalization with latent structure and data augmentation}.
\newblock In \emph{Proceedings of the 2022 Conference of the North American
  Chapter of the Association for Computational Linguistics: Human Language
  Technologies}, pages 4341--4362, Seattle, United States. Association for
  Computational Linguistics.

\bibitem[{Rae et~al.(2021)Rae, Borgeaud, Cai, Millican, Hoffmann, Song,
  Aslanides, Henderson, Ring, Young, Rutherford, Hennigan, Menick, Cassirer,
  Powell, van~den Driessche, Hendricks, Rauh, Huang, Glaese, Welbl, Dathathri,
  Huang, Uesato, Mellor, Higgins, Creswell, McAleese, Wu, Elsen, Jayakumar,
  Buchatskaya, Budden, Sutherland, Simonyan, Paganini, Sifre, Martens, Li,
  Kuncoro, Nematzadeh, Gribovskaya, Donato, Lazaridou, Mensch, Lespiau,
  Tsimpoukelli, Grigorev, Fritz, Sottiaux, Pajarskas, Pohlen, Gong, Toyama,
  de~Masson~d'Autume, Li, Terzi, Mikulik, Babuschkin, Clark, de~Las~Casas, Guy,
  Jones, Bradbury, Johnson, Hechtman, Weidinger, Gabriel, Isaac, Lockhart,
  Osindero, Rimell, Dyer, Vinyals, Ayoub, Stanway, Bennett, Hassabis,
  Kavukcuoglu, and Irving}]{rae2021scaling}
Jack~W. Rae, Sebastian Borgeaud, Trevor Cai, Katie Millican, Jordan Hoffmann,
  Francis Song, John Aslanides, Sarah Henderson, Roman Ring, Susannah Young,
  Eliza Rutherford, Tom Hennigan, Jacob Menick, Albin Cassirer, Richard Powell,
  George van~den Driessche, Lisa~Anne Hendricks, Maribeth Rauh, Po-Sen Huang,
  Amelia Glaese, Johannes Welbl, Sumanth Dathathri, Saffron Huang, Jonathan
  Uesato, John Mellor, Irina Higgins, Antonia Creswell, Nat McAleese, Amy Wu,
  Erich Elsen, Siddhant Jayakumar, Elena Buchatskaya, David Budden, Esme
  Sutherland, Karen Simonyan, Michela Paganini, Laurent Sifre, Lena Martens,
  Xiang~Lorraine Li, Adhiguna Kuncoro, Aida Nematzadeh, Elena Gribovskaya,
  Domenic Donato, Angeliki Lazaridou, Arthur Mensch, Jean-Baptiste Lespiau,
  Maria Tsimpoukelli, Nikolai Grigorev, Doug Fritz, Thibault Sottiaux, Mantas
  Pajarskas, Toby Pohlen, Zhitao Gong, Daniel Toyama, Cyprien
  de~Masson~d'Autume, Yujia Li, Tayfun Terzi, Vladimir Mikulik, Igor
  Babuschkin, Aidan Clark, Diego de~Las~Casas, Aurelia Guy, Chris Jones, James
  Bradbury, Matthew Johnson, Blake Hechtman, Laura Weidinger, Iason Gabriel,
  William Isaac, Ed~Lockhart, Simon Osindero, Laura Rimell, Chris Dyer, Oriol
  Vinyals, Kareem Ayoub, Jeff Stanway, Lorrayne Bennett, Demis Hassabis, Koray
  Kavukcuoglu, and Geoffrey Irving. 2021.
\newblock \href {http://arxiv.org/abs/2112.11446} {Scaling language models:
  Methods, analysis \& insights from training gopher}.
\newblock \emph{CoRR}.

\bibitem[{Raffel et~al.(2020)Raffel, Shazeer, Roberts, Lee, Narang, Matena,
  Zhou, Li, and Liu}]{raffel2019exploring}
Colin Raffel, Noam Shazeer, Adam Roberts, Katherine Lee, Sharan Narang, Michael
  Matena, Yanqi Zhou, Wei Li, and Peter~J Liu. 2020.
\newblock Exploring the limits of transfer learning with a unified text-to-text
  transformer.
\newblock \emph{Journal of Machine Learning Research}, 21:1--67.

\bibitem[{Rajkumar et~al.(2022)Rajkumar, Li, and
  Bahdanau}]{rajkumar2022evaluating}
Nitarshan Rajkumar, Raymond Li, and Dzmitry Bahdanau. 2022.
\newblock \href {https://arxiv.org/abs/2204.00498} {Evaluating the text-to-sql
  capabilities of large language models}.
\newblock \emph{ArXiv preprint}, abs/2204.00498.

\bibitem[{Reynolds and McDonell(2021)}]{reynolds2021prompt}
Laria Reynolds and Kyle McDonell. 2021.
\newblock \href {https://doi.org/10.1145/3411763.3451760} {Prompt programming
  for large language models: Beyond the few-shot paradigm}.
\newblock In \emph{{CHI} '21: {CHI} Conference on Human Factors in Computing
  Systems, Virtual Event / Yokohama Japan, May 8-13, 2021, Extended Abstracts},
  pages 314:1--314:7. {ACM}.

\bibitem[{Robertson and Zaragoza(2009)}]{robertson2009probabilistic}
Stephen~E. Robertson and Hugo Zaragoza. 2009.
\newblock \href {https://doi.org/10.1561/1500000019} {The probabilistic
  relevance framework: {BM25} and beyond}.
\newblock \emph{Found. Trends Inf. Retr.}, 3(4):333--389.

\bibitem[{Rosenfeld et~al.(2020)Rosenfeld, Rosenfeld, Belinkov, and
  Shavit}]{rosenfeld2020constructive}
Jonathan~S. Rosenfeld, Amir Rosenfeld, Yonatan Belinkov, and Nir Shavit. 2020.
\newblock \href {https://openreview.net/forum?id=ryenvpEKDr} {A constructive
  prediction of the generalization error across scales}.
\newblock In \emph{8th International Conference on Learning Representations,
  {ICLR} 2020, Addis Ababa, Ethiopia, April 26-30, 2020}. OpenReview.net.

\bibitem[{Rubin et~al.(2021)Rubin, Herzig, and
  Berant}]{rubin-etal-2022-learning}
Ohad Rubin, Jonathan Herzig, and Jonathan Berant. 2021.
\newblock \href {https://arxiv.org/abs/2112.08633} {Learning to retrieve
  prompts for in-context learning}.
\newblock \emph{ArXiv preprint}, abs/2112.08633.

\bibitem[{Ruiz et~al.(2021)Ruiz, Ainslie, and
  Onta{\~n}{\'o}n}]{ruiz2021iterative}
Luana Ruiz, Joshua Ainslie, and Santiago Onta{\~n}{\'o}n. 2021.
\newblock \href {https://arxiv.org/abs/2110.04169} {Iterative decoding for
  compositional generalization in transformers}.
\newblock \emph{ArXiv preprint}, abs/2110.04169.

\bibitem[{Russin et~al.(2019)Russin, Jo, O'Reilly, and
  Bengio}]{russin2019compositional}
Jake Russin, Jason Jo, Randall~C O'Reilly, and Yoshua Bengio. 2019.
\newblock \href {https://arxiv.org/abs/1904.09708} {Compositional
  generalization in a deep seq2seq model by separating syntax and semantics}.
\newblock \emph{ArXiv preprint}, abs/1904.09708.

\bibitem[{Scholak et~al.(2021)Scholak, Schucher, and
  Bahdanau}]{scholak-etal-2021-picard}
Torsten Scholak, Nathan Schucher, and Dzmitry Bahdanau. 2021.
\newblock \href {https://doi.org/10.18653/v1/2021.emnlp-main.779} {{PICARD}:
  Parsing incrementally for constrained auto-regressive decoding from language
  models}.
\newblock In \emph{Proceedings of the 2021 Conference on Empirical Methods in
  Natural Language Processing}, pages 9895--9901, Online and Punta Cana,
  Dominican Republic. Association for Computational Linguistics.

\bibitem[{Schucher et~al.(2022)Schucher, Reddy, and
  de~Vries}]{schucher-etal-2022-power}
Nathan Schucher, Siva Reddy, and Harm de~Vries. 2022.
\newblock \href {https://doi.org/10.18653/v1/2022.acl-short.17} {The power of
  prompt tuning for low-resource semantic parsing}.
\newblock In \emph{Proceedings of the 60th Annual Meeting of the Association
  for Computational Linguistics (Volume 2: Short Papers)}, pages 148--156,
  Dublin, Ireland. Association for Computational Linguistics.

\bibitem[{Shaw et~al.(2021)Shaw, Chang, Pasupat, and
  Toutanova}]{shaw-etal-2021-compositional}
Peter Shaw, Ming-Wei Chang, Panupong Pasupat, and Kristina Toutanova. 2021.
\newblock \href {https://doi.org/10.18653/v1/2021.acl-long.75} {Compositional
  generalization and natural language variation: Can a semantic parsing
  approach handle both?}
\newblock In \emph{Proceedings of the 59th Annual Meeting of the Association
  for Computational Linguistics and the 11th International Joint Conference on
  Natural Language Processing (Volume 1: Long Papers)}, pages 922--938, Online.
  Association for Computational Linguistics.

\bibitem[{Shin and Durme(2021)}]{shin2021few}
Richard Shin and Benjamin~Van Durme. 2021.
\newblock \href {https://arxiv.org/abs/2112.08696} {Few-shot semantic parsing
  with language models trained on code}.
\newblock \emph{ArXiv preprint}, abs/2112.08696.

\bibitem[{Shin et~al.(2021)Shin, Lin, Thomson, Chen, Roy, Platanios, Pauls,
  Klein, Eisner, and Van~Durme}]{shin-etal-2021-constrained}
Richard Shin, Christopher Lin, Sam Thomson, Charles Chen, Subhro Roy,
  Emmanouil~Antonios Platanios, Adam Pauls, Dan Klein, Jason Eisner, and
  Benjamin Van~Durme. 2021.
\newblock \href {https://doi.org/10.18653/v1/2021.emnlp-main.608} {Constrained
  language models yield few-shot semantic parsers}.
\newblock In \emph{Proceedings of the 2021 Conference on Empirical Methods in
  Natural Language Processing}, pages 7699--7715, Online and Punta Cana,
  Dominican Republic. Association for Computational Linguistics.

\bibitem[{Tang and Mooney(2001)}]{tang2001using}
Lappoon~R Tang and Raymond~J Mooney. 2001.
\newblock Using multiple clause constructors in inductive logic programming for
  semantic parsing.
\newblock In \emph{European Conference on Machine Learning}, pages 466--477.
  Springer.

\bibitem[{Tay et~al.(2021)Tay, Dehghani, Rao, Fedus, Abnar, Chung, Narang,
  Yogatama, Vaswani, and Metzler}]{tay2021scale}
Yi~Tay, Mostafa Dehghani, Jinfeng Rao, William Fedus, Samira Abnar, Hyung~Won
  Chung, Sharan Narang, Dani Yogatama, Ashish Vaswani, and Donald Metzler.
  2021.
\newblock \href {https://arxiv.org/abs/2109.10686} {Scale efficiently: Insights
  from pre-training and fine-tuning transformers}.
\newblock \emph{ArXiv preprint}, abs/2109.10686.

\bibitem[{Tsarkov et~al.(2021)Tsarkov, Tihon, Scales, Momchev, Sinopalnikov,
  and Sch{\"{a}}rli}]{tsarkov2021cfq}
Dmitry Tsarkov, Tibor Tihon, Nathan Scales, Nikola Momchev, Danila
  Sinopalnikov, and Nathanael Sch{\"{a}}rli. 2021.
\newblock \href {https://ojs.aaai.org/index.php/AAAI/article/view/17195}
  {*-cfq: Analyzing the scalability of machine learning on a compositional
  task}.
\newblock In \emph{Thirty-Fifth {AAAI} Conference on Artificial Intelligence,
  {AAAI} 2021, Thirty-Third Conference on Innovative Applications of Artificial
  Intelligence, {IAAI} 2021, The Eleventh Symposium on Educational Advances in
  Artificial Intelligence, {EAAI} 2021, Virtual Event, February 2-9, 2021},
  pages 9949--9957. {AAAI} Press.

\bibitem[{Wang et~al.(2021)Wang, Lapata, and Titov}]{wang2021structured}
Bailan Wang, Mirella Lapata, and Ivan Titov. 2021.
\newblock Structured reordering for modeling latent alignments in sequence
  transduction.
\newblock \emph{Advances in Neural Information Processing Systems}, 34.

\bibitem[{Wang et~al.(2022)Wang, Roberts, Hesslow, Scao, Chung, Beltagy,
  Launay, and Raffel}]{wang2022what}
Thomas Wang, Adam Roberts, Daniel Hesslow, Teven~Le Scao, Hyung~Won Chung,
  Iz~Beltagy, Julien Launay, and Colin Raffel. 2022.
\newblock \href {https://doi.org/10.48550/arXiv.2204.05832} {What language
  model architecture and pretraining objective work best for zero-shot
  generalization?}
\newblock \emph{CoRR}, abs/2204.05832.

\bibitem[{Wortsman et~al.(2021)Wortsman, Ilharco, Li, Kim, Hajishirzi, Farhadi,
  Namkoong, and Schmidt}]{wortsman2021robust}
Mitchell Wortsman, Gabriel Ilharco, Mike Li, Jong~Wook Kim, Hannaneh
  Hajishirzi, Ali Farhadi, Hongseok Namkoong, and Ludwig Schmidt. 2021.
\newblock \href {https://arxiv.org/abs/2109.01903} {Robust fine-tuning of
  zero-shot models}.
\newblock \emph{ArXiv preprint}, abs/2109.01903.

\bibitem[{Xie et~al.(2022)Xie, Wu, Shi, Zhong, Scholak, Yasunaga, Wu, Zhong,
  Yin, Wang, Zhong, Wang, Li, Boyle, Ni, Yao, Radev, Xiong, Kong, Zhang, Smith,
  Zettlemoyer, and Yu}]{xie2022unifiedskg}
Tianbao Xie, Chen~Henry Wu, Peng Shi, Ruiqi Zhong, Torsten Scholak, Michihiro
  Yasunaga, Chien{-}Sheng Wu, Ming Zhong, Pengcheng Yin, Sida~I. Wang, Victor
  Zhong, Bailin Wang, Chengzu Li, Connor Boyle, Ansong Ni, Ziyu Yao,
  Dragomir~R. Radev, Caiming Xiong, Lingpeng Kong, Rui Zhang, Noah~A. Smith,
  Luke Zettlemoyer, and Tao Yu. 2022.
\newblock \href {https://arxiv.org/abs/2201.05966} {Unifiedskg: Unifying and
  multi-tasking structured knowledge grounding with text-to-text language
  models}.
\newblock \emph{ArXiv preprint}, abs/2201.05966.

\bibitem[{Yang et~al.(2022)Yang, Zhang, and Yang}]{yang2022subs}
Jingfeng Yang, Le~Zhang, and Diyi Yang. 2022.
\newblock \href {https://doi.org/10.48550/arXiv.2205.01538} {{SUBS:} subtree
  substitution for compositional semantic parsing}.
\newblock \emph{CoRR}, abs/2205.01538.

\bibitem[{Yin et~al.(2021)Yin, Fang, Neubig, Pauls, Platanios, Su, Thomson, and
  Andreas}]{yin2021compositional}
Pengcheng Yin, Hao Fang, Graham Neubig, Adam Pauls, Emmanouil~Antonios
  Platanios, Yu~Su, Sam Thomson, and Jacob Andreas. 2021.
\newblock \href {https://doi.org/10.18653/v1/2021.naacl-main.225}
  {Compositional generalization for neural semantic parsing via span-level
  supervised attention}.
\newblock In \emph{Proceedings of the 2021 Conference of the North American
  Chapter of the Association for Computational Linguistics: Human Language
  Technologies}, pages 2810--2823, Online. Association for Computational
  Linguistics.

\bibitem[{Zelle and Mooney(1996)}]{zelle1996learning}
John~M Zelle and Raymond~J Mooney. 1996.
\newblock Learning to parse database queries using inductive logic programming.
\newblock In \emph{Proceedings of the thirteenth national conference on
  Artificial intelligence-Volume 2}, pages 1050--1055.

\bibitem[{Zhang et~al.(2022)Zhang, Roller, Goyal, Artetxe, Chen, Chen, Dewan,
  Diab, Li, Lin, Mihaylov, Ott, Shleifer, Shuster, Simig, Koura, Sridhar, Wang,
  and Zettlemoyer}]{zhang2022opt}
Susan Zhang, Stephen Roller, Naman Goyal, Mikel Artetxe, Moya Chen, Shuohui
  Chen, Christopher Dewan, Mona Diab, Xian Li, Xi~Victoria Lin, Todor Mihaylov,
  Myle Ott, Sam Shleifer, Kurt Shuster, Daniel Simig, Punit~Singh Koura, Anjali
  Sridhar, Tianlu Wang, and Luke Zettlemoyer. 2022.
\newblock \href {https://doi.org/10.48550/arXiv.2205.01068} {{OPT:} open
  pre-trained transformer language models}.
\newblock \emph{CoRR}, abs/2205.01068.

\bibitem[{Zhao et~al.(2021)Zhao, Wallace, Feng, Klein, and
  Singh}]{zhao2021calibrating}
Zihao Zhao, Eric Wallace, Shi Feng, Dan Klein, and Sameer Singh. 2021.
\newblock \href {http://proceedings.mlr.press/v139/zhao21c.html} {Calibrate
  before use: Improving few-shot performance of language models}.
\newblock In \emph{Proceedings of the 38th International Conference on Machine
  Learning, {ICML} 2021, 18-24 July 2021, Virtual Event}, volume 139 of
  \emph{Proceedings of Machine Learning Research}, pages 12697--12706. {PMLR}.

\bibitem[{Zheng and Lapata(2021)}]{zheng2020compositional}
Hao Zheng and Mirella Lapata. 2021.
\newblock \href {https://doi.org/10.18653/v1/2021.findings-emnlp.88}
  {Compositional generalization via semantic tagging}.
\newblock In \emph{Findings of the Association for Computational Linguistics:
  EMNLP 2021}, pages 1022--1032, Punta Cana, Dominican Republic. Association
  for Computational Linguistics.

\bibitem[{Zhu et~al.(2021)Zhu, Shaw, Linzen, and Sha}]{zhu2021learning}
Wang Zhu, Peter Shaw, Tal Linzen, and Fei Sha. 2021.
\newblock \href {https://arxiv.org/abs/2111.05013} {Learning to generalize
  compositionally by transferring across semantic parsing tasks}.
\newblock \emph{ArXiv preprint}, abs/2111.05013.

\end{thebibliography}
\bibliographystyle{acl_natbib}

\clearpage
\appendix
\section*{Appendix}
The appendix is organized into three sections: 
dataset details (Appendix~\ref{sec:appendix_dataset}), experiment details (Appendix~\ref{sec:appendix_experiment}), and additional results and analysis (Appendix~\ref{sec:appendix_results}).

\section{Dataset Details}
\label{sec:appendix_dataset}

\subsection{Dataset Sizes}
We follow prior work~\citep{qiu2022improving} and use the same splits for GeoQuery. We evaluate on the small subset of COGS~\cite{kim-linzen-2020-cogs} and CFQ~\cite{keysers2019measuring}. We use the newer version of SMCalFlow and re-ran the data generation pipeline from~\citet{yin2021compositional} to create SMCalFlow-CS. Dataset sizes are shown in Table~\ref{tab:dataset_sizes}.

\begin{table}[h!]
\centering
\resizebox{0.95\columnwidth}{!}{
\begin{tabular}{@{}llccc@{}}
\toprule
Dataset                       & Split     & Train & Dev & Test \\
\midrule   
\multirow{2}{*}{COGS}         & In-dist.        & 24K   & 1000 & 1000  \\
                              & Gen.      & 24K   & 1050 & 1050  \\
\midrule
\multirow{4}{*}{CFQ}          & Random    & 95743 & 1000 & 1000 \\
                              & MCD1      & 95743 & 1000 & 1000 \\
                              & MCD2      & 95743 & 1000 & 1000 \\
                              & MCD3      & 95743 & 1000 & 1000 \\
\midrule      
\multirow{8}{*}{GeoQuery}     & Standard  & 600 & --- & 280 \\
                              & Template1 & 438 & 110 & 332 \\
                              & Template2 & 439 & 110 & 331 \\
                              & Template3 & 440 & 110 & 330 \\
                              & TMCD1     & 440 & 110 & 330 \\
                              & TMCD2     & 440 & 110 & 330 \\
                              & TMCD3     & 440 & 110 & 330 \\
                              & Length    & 440 & 110 & 330 \\
\midrule
\multirow{4}{*}{SMCalFlow-CS} & 8-shot    & 20965 & 360 & 360 \\
                              & 16-shot   & 20973 & 360 & 360 \\
                              & 32-shot   & 20989 & 360 & 360 \\
                              & Length    & 20237 & 360 & 360 \\
                        
\bottomrule
\end{tabular}}
\caption{Sizes of all datasets and splits.}
\label{tab:dataset_sizes}
\end{table}

\subsection{Intermediate Representation}
\label{sec:appendix_ir}
We consider different output formats for COGS and CFQ. For COGS, we choose the variable-free form used in~\citet{qiu2022improving} as intermediate representation. For CFQ, we use the reversible intermediate representation in~\citet{herzig2021unlocking}. Table~\ref{tab:output_space} shows examples of input-output pairs. 
\begin{table*}[t!]
  \begin{center}
  \scalebox{0.82}{
  \begin{tabular}{ll}
  \toprule
  $x$: & Camila gave a cake in a storage to Emma . \\
  $y$: & \texttt{give . agent ( x$\_$1 , Camila ) AND give . theme ( x$\_$1 , x$\_$3 )} \\
  & \texttt{AND give . recipient ( x$\_$1 , Emma ) AND cake ( x$\_$3 )} \\
  & \texttt{AND cake . nmod . in ( x$\_$3 , x$\_$6 ) AND storage ( x$\_$6 )} \\
  $y'$: & \texttt{give ( agent = Camila , theme = cake ( nmod . in = storage , recipient = Emma )} \\
  \midrule
  $x$: & Did a film 's editor executive produce , write , and direct M0 , M1 , and M2 \\
  $y$: & \texttt{SELECT count(*) WHERE $\{$ ?x0 ns:film.director.film M0 . } \\
  & \texttt{?x0 ns:film.director.film M1 . ?x0 ns:film.director.film M2 . } \\
  & \texttt{?x0 ns:film.editor.film ?x1 . ?x0 ns:film.producer.films$\_$executive$\_$produced M0 . } \\
  & \texttt{?x0 ns:film.producer.films$\_$executive$\_$produced M1 . } \\
  & \texttt{?x0 ns:film.producer.films$\_$executive$\_$produced M2 . ?x0 ns:film.writer.film M0 . } \\
  & \texttt{?x0 ns:film.writer.film M1 . ?x0 ns:film.writer.film M2 . ?x1 a ns:film.film $\}$ } \\
  $y'$: & \texttt{SELECT count(*) WHERE $\{$ ( ?x0 ( film.director.film , } \\
  & \texttt{film.producer.films$\_$executive$\_$produced , film.writer.film ) ( M0 , M1 , M2 ) ) .} \\
  & \texttt{( ?x0 ( film.editor.film ) ( ?x1 ) ) . ( ?x1 a film.film ) $\}$ } \\
  \bottomrule
  \end{tabular}
  }
  \end{center}
  \caption{An example input $x$, output $y$, and intermediate representation $y'$ from COGS (top) and CFQ (bottom).}
  \label{tab:output_space}
  \end{table*}

\section{Experiment Details}
\label{sec:appendix_experiment}

\subsection{Experimental Setup}
\label{sec:appendix_experimental_setup}
\paragraph{Training}
For fine-tuning, we use learning rate of $1e^{-4}$ for GeoQuery and COGS and $1e^{-3}$ for CFQ and SMCalFlow-CS. We select learning rate from $[1e^{-3}, 1e^{-4}, 1e^{-5}]$ based on validation accuracy. For prompt tuning, we use learning rate of $0.3$ for GeoQuery and COGS and $1.0$ for CFQ and SMCalFlow-CS. The learning rate is selected from $[0.3, 1, 3]$. We use a tunable prompt length of $100$ for all prompt tuning experiments. In-context learning does not require any training. We only tune hyperparameters using smaller models (T5-base and PaLM-8B for fine-tuning, T5-3B for prompt tuning) to optimize computational resources. Tuning hyperparameters for each model scale could potentially further improve performance, but is not the focus of our study. We train all models on Cloud TPU. The training time varies across different datasets and model sizes. The shortest training takes around 1 hour and the longest training takes around 5 days.  

\paragraph{Inference} For fine-tuning and prompt tuning, we only use the test query as input. For in-context learning, we retrieve $K$ exemplars from the training set and concatenate each exemplar to the query. We add special prefixes ``In: '' and ``Out: '' for retrieved input-output pairs and separate exemplars with break lines. We sort exemplars based on their similarities to the query in ascending order. Empirically, we find putting the most similar exemplar close to the query works better than the reverse. We use the maximum number of exemplars up to 1,920 tokens. We use greedy decoding for all models.

\subsection{Number of Exemplars}
We use the maximum number of exemplars up to 1,920 tokens for all in-context learning experiments. We show the mean and standard deviation of number of exemplars for each split in Table~\ref{tab:num_exemplars}.

\begin{table}[!t]
\begin{center}

\scalebox{0.76}{
\begin{tabular}{llcccc}
\toprule
 \multirow{2}{*}{Dataset} & \multirow{2}{*}{Split} & \multicolumn{2}{c}{Non-oracle}
 & \multicolumn{2}{c}{Oracle} \\
 \cmidrule(lr){3-4} \cmidrule(lr){5-6}
 & & Mean & Stdev. & Mean & Stdev. \\
\midrule
\multirow{2}{*}{COGS}     & In-dist.         & 58.1 & 7.2  & 59.1 & 6.1    \\
                          & Gen.       & 54.1 & 8.8  & 57.1 & 7.7    \\
\midrule
\multirow{4}{*}{CFQ}      & Random & 12.1 & 3.3 & 12.8 & 4.7 \\
                          & MCD1 & 12.4 & 4.0 & 16.2 & 4.8 \\
                          & MCD2 & 12.4 & 4.2 & 14.9 & 4.7 \\
                          & MCD3 & 15.1 & 5.1 & 15.5 & 4.7 \\
\midrule
\multirow{8}{*}{GeoQuery} & Std.       & 44.1 & 4.0  & 48.8 & 8.2    \\
                          & Template1  & 43.7 & 2.8  & 47.5 & 8.5    \\
                          & Template2  & 42.8 & 3.3  & 42.7 & 5.5    \\
                          & Template3  & 43.8 & 3.8  & 47.3 & 7.6    \\
                          & TMCD1      & 44.1 & 3.4  & 41.9 & 4.7    \\
                          & TMCD2      & 46.6 & 2.9  & 49.9 & 6.8    \\
                          & TMCD3      & 43.7 & 3.5  & 49.5 & 7.8    \\
                          & Length     & 59.3 & 3.3  & 53.3 & 3.1    \\
\midrule
\multirow{7}{*}{SMCalFlow-CS} & 8-S    & 24.0 & 5.6  & 26.2 & 9.7    \\
                              & 8-C    & 22.9 & 6.5  & 22.3 & 10.0   \\
                              & 16-S   & 24.0 & 5.6  & 26.2 & 9.7    \\
                              & 16-C   & 22.8 & 6.3  & 22.2 & 9.6    \\
                              & 32-S   & 23.9 & 5.6  & 26.2 & 9.7    \\
                              & 32-C   & 22.2 & 6.2  & 22.0 & 8.9    \\
                              & Length & 19.3 & 3.9  & 16.7 & 5.3    \\
\bottomrule
\end{tabular}
}
\caption{The mean and standard deviation of number of exemplars for in-context learning.}
\label{tab:num_exemplars}
\end{center}
\end{table}
\section{Additional Results}
\label{sec:appendix_results}

\begin{table*}[!t]
\begin{center}
\scalebox{0.82}{
\begin{tabular}{lccccccccccc}
\toprule
 & \multicolumn{2}{c}{{\bf{\textsc{COGS}}}} 
 & \multicolumn{2}{c}{{\bf{\textsc{CFQ}}}} 
 & \multicolumn{4}{c}{{\bf{\textsc{GeoQuery}}}} 
 & \multicolumn{3}{c}{{\bf\textsc{SMCalFlow-CS}}} \\
 \cmidrule(lr){2-3} \cmidrule(lr){4-5} \cmidrule(lr){6-9} \cmidrule(lr){10-12}
 & In-dist. & Gen. & Random & MCD & Std. & Templ. & TMCD & Len. & Single & Cross & Len. \\
\midrule
BERT Coverage & 97.8 & 67.0 & 85.6 & 65.0 & 99.3 & 94.2 & 88.5 & 90.9 & 75.8 & 7.3 & 36.1 \\
BERT Precision & 86.6 & 65.3 & 65.9 & 11.6 & 87.1 & 73.2 & 69.8 & 54.0 & 67.2 & 18.3 & 16.2 \\
BM25 Coverage & 100.0 & 97.3 & 93.4 & 76.9 & 99.6 & 96.1 & 96.4 & 94.5 & 92.2 & 57.7 & 68.3 \\
BM25 Precision & 77.1 & 60.3 & 58.0 & 10.5 & 87.1 & 76.3 & 68.7 & 55.8 & 64.5 & 16.3 & 22.4 \\
\midrule
BM25 Oracle Coverage & 100.0 & 100.0 & 99.8 & 99.8 & 99.6 & 97.9 & 100.0 & 96.4 & 99.7 & 100.0 & 96.9\\
BM25 Oracle Precision & 77.0 & 58.0 & 58.1 & 14.8 & 91.0 & 73.7 & 67.9 & 57.5 & 73.6 & 30.0 & 35.5 \\
Target Overlap Coverage & 100.0 & 100.0 & 100.0 & 99.9 & 99.6 & 97.6 & 100.0 & 94.5 & 98.1 & 99.7 & 91.4 \\
Target Overlap Precision & 91.0 & 69.0 & 74.6 & 21.7 & 92.5 & 76.4 & 75.5 & 59.6 & 87.3 & 43.6 & 49.8 \\
\bottomrule
\end{tabular}
}
\caption{We compare the coverage and precision of different types of retrievers on development set using PaLM-540B model. We show results of non-oracle retrievers (top) and oracle retrievers (bottom).}
\label{tab:retriever_breakdown}
\end{center}
\end{table*}

\subsection{Additional Retriever Analysis} 
\label{sec:appendix_retriever_analysis}

\paragraph{Token Coverage and Precision}
We compute the coverage (fraction of examples where the exemplars and test query contain all tokens of gold target) and precision (fraction of examples with a correct prediction among ones where the example is covered) of different retrievers. Note that the actual accuracy can be higher than the product of coverage and precision, as models can generate correct outputs even without full coverage of tokens. 

The results are shown in Table~\ref{tab:retriever_breakdown}. For intra-class comparison, the non-oracle retriever BM25 has higher coverage but lower precision compared to BERT, as the TF-IDF based score allows retrieving exemplars containing symbols with lower frequency. This is more desirable for compositional splits, leading to larger performance improvement using BM25 than BERT on most splits, except for COGS.
For oracle retrievers, retrieving exemplars based on the overlap of target compounds outperforms using BM25 on the target itself. 
This suggests the importance of considering larger output structures as opposed to single tokens. Note that the target overlap oracle does not consider the order of sub-structures. Modeling this could potentially lead to further improvements.

For inter-class comparison, the oracle retrievers generally have better accuracy and coverage than their non-oracle counterparts. They also have better precision within examples that are covered, suggesting atom coverage is not the only factor to consider when designing retrievers. For non-synthetic splits with large training sets like SMCalFlow-CS, we observe significant performance gain when switching from a non-oracle retriever to an oracle retriever, which suggests that improving retrieval for in-context learning is an important direction for future work.

\subsection{Results on Individual Splits}
\label{sec:individual-split-results}
\begin{figure*}[ht!]
\centering
\includegraphics[width=\linewidth]{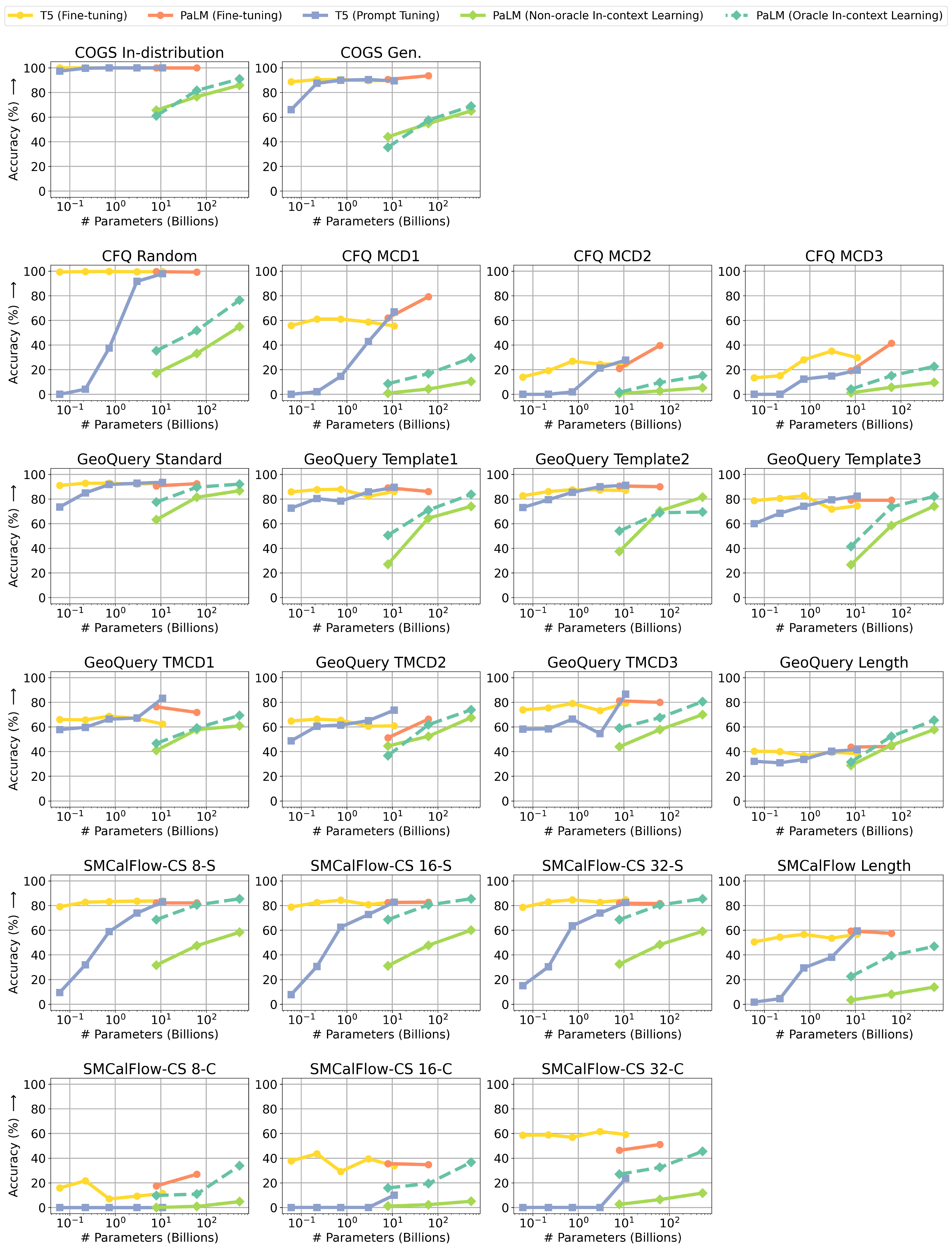}
\caption{Scaling curves for individual split of different datasets. Note that the in-context learning with an oracle retriever (dashed) cannot be compared directly with other methods as it has access to the gold output.}
\label{fig:all_scale}
\end{figure*}

\begin{figure*}[ht!]
\centering
\includegraphics[width=\linewidth]{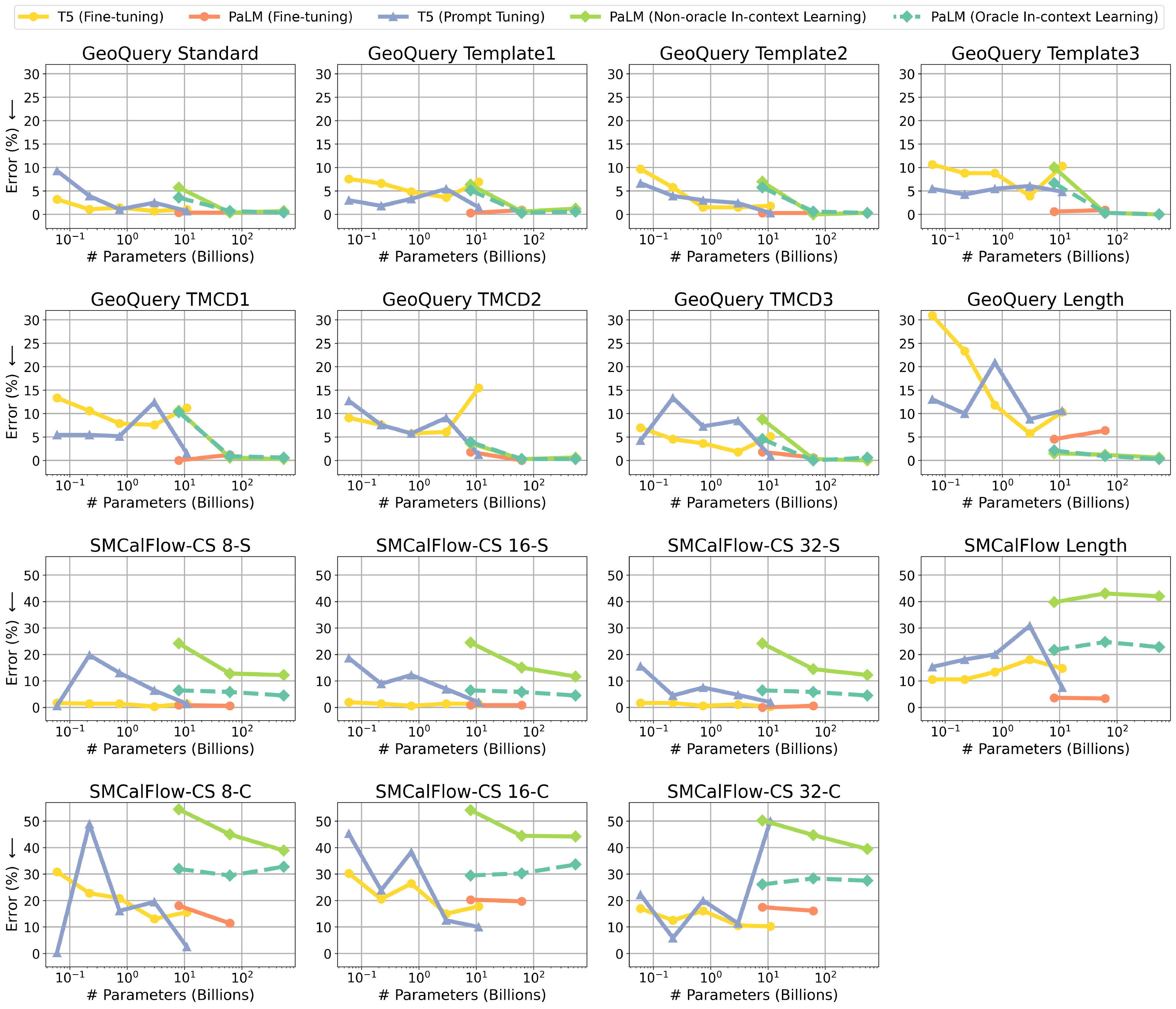}
\caption{Percentage of predictions that contain unbalanced parentheses, as an estimate of syntax errors.}
\label{fig:all_syntax_error}
\end{figure*}

\begin{figure*}[ht!]
\centering
\includegraphics[width=\linewidth]{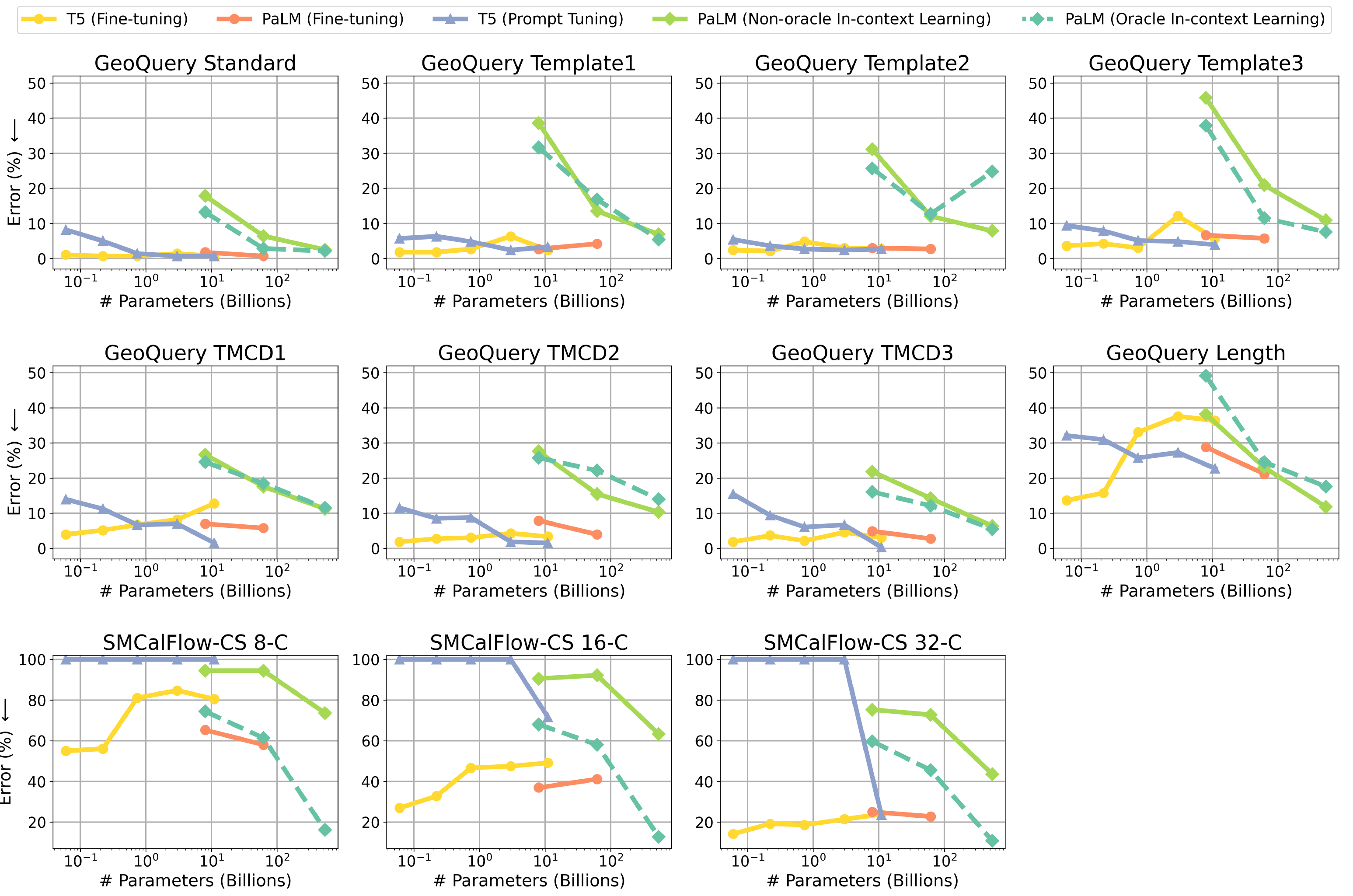}
\caption{Percentage of incorrect predictions where the output exactly matches an output seen in the training set on GeoQuery splits (top two rows). Percentage of errors where the prediction does not contain cross-domain predicates on SMCalFlow-CS cross-domain splits (bottom).}
\label{fig:all_compositional_error}
\end{figure*}

We show full fine-tuning results in Table~\ref{tab:ft_results} and full prompt tuning and in-context learning results in Table~\ref{tab:pt_icl_results}.
We include scaling curves of individual splits in Figure~\ref{fig:all_scale}. We also show error trends of individual non-synthetic splits in Figure~\ref{fig:all_syntax_error} and Figure~\ref{fig:all_compositional_error}.

\begin{table*}[!ht]
\centering
\scalebox{0.82}{
\begin{tabular}{llccccccc}
\toprule
 & 
 & \multicolumn{5}{c}{T5 (FT)} 
 & \multicolumn{2}{c}{PaLM (FT)} \\
 \cmidrule(lr){3-7} \cmidrule(lr){8-9} 
Dataset                       & Split & Small & Base & Large & 3B & 11B & 8B & 62B \\
\midrule
\multirow{2}{*}{COGS}         & In-dist. & 100.0 & 100.0 & 100.0 & 100.0 & 100.0 & 100.0 & 100.0 \\
                              & Gen. & 88.7 & 90.5 & 90.5 & 89.6 & 89.8 & 90.6 & 93.6 \\
\midrule
\multirow{4}{*}{CFQ}          & Random & 99.4 & 99.6 & 99.7 & 99.5 & 99.6 & 99.6 & 99.2 \\
                              & MCD1 & 55.9 & 61.1 & 61.1 & 58.7 & 55.5 & 62.0 & 79.2 \\
                              & MCD2 & 14.0 & 19.2 & 26.9 & 24.5 & 24.7 & 21.0 & 39.7 \\
                              & MCD3 & 13.4 & 15.1 & 28.1 & 35.1 & 29.8 & 19.1 & 41.4 \\
\midrule      
\multirow{8}{*}{GeoQuery}     & Std. & 91.1 & 92.9 & 92.9 & 92.1 & 92.9 & 90.7 & 92.5 \\
                              & Template1 & 85.8 & 87.7 & 88.0 & 82.2 & 86.1 & 89.2 & 86.1 \\
                              & Template2 & 82.8 & 86.1 & 87.6 & 87.3 & 87.0 & 90.6 & 90.0 \\
                              & Template3 & 78.8 & 80.6 & 82.7 & 71.8 & 74.5 & 79.1 & 79.1 \\
                              & TMCD1 & 66.1 & 65.8 & 68.8 & 67.0 & 62.4 & 76.4 & 71.8 \\
                              & TMCD2 & 64.8 & 66.4 & 65.5 & 60.6 & 60.9 & 51.2 & 66.4 \\
                              & TMCD3 & 73.9 & 75.5 & 79.1 & 73.3 & 79.4 & 81.2 & 80.0 \\
                              & Length & 40.3 & 40.0 & 36.7 & 39.4 & 38.5 & 43.6 & 44.2 \\
\midrule
\multirow{7}{*}{SMCalFlow-CS} & 8-S & 79.2 & 82.8 & 83.3 & 83.6 & 83.9 & 82.2 & 82.2 \\
                              & 8-C & 15.8 & 21.7 & 6.9 & 9.2 & 11.4 & 17.5 & 26.9 \\
                              & 16-S & 78.9 & 82.5 & 84.4 & 80.8 & 83.1 & 82.5 & 82.8 \\
                              & 16-C & 37.8 & 43.6 & 29.2 & 39.4 & 33.9 & 35.6 & 34.7 \\
                              & 32-S & 78.6 & 83.1 & 84.7 & 82.8 & 84.7 & 81.9 & 81.7 \\
                              & 32-C & 58.6 & 58.9 & 56.9 & 61.7 & 59.2 & 46.4 & 51.1 \\
                              & Length & 50.6 & 54.4 & 56.7 & 53.6 & 56.7 & 59.4 & 57.5 \\
\bottomrule
\end{tabular}}
\caption{Fine-tuning (FT) results on all datasets and splits.}
\label{tab:ft_results}
\end{table*}

\begin{table*}[!ht]
\centering
\scalebox{0.8}{
\begin{tabular}{p{24mm}p{15mm}*{11}{>{\centering\arraybackslash}p{9.3mm}}}
\toprule
 & 
 & \multicolumn{5}{c}{T5 (PT)} 
 & \multicolumn{3}{c}{PaLM (Non-oracle ICL)}
 & \multicolumn{3}{c}{PaLM (Oracle ICL)} \\
 \cmidrule(lr){3-7} \cmidrule(lr){8-10} \cmidrule(lr){11-13} 
Dataset                       & Split & Small & Base & Large & 3B & 11B & 8B & 62B & 540B & 8B & 62B & 540B \\
\midrule
\multirow{2}{*}{COGS}         & In-dist. & 97.4 & 99.7 & 100.0 & 100.0 & 100.0 & 44.8 & 63.2 & 77.1 & 61.1 & 81.7 & 91.0 \\
                              & Gen. & 66.1 & 87.5 & 90.0 & 90.5 & 89.5 & 32.1 & 44.8 & 58.8 & 35.5 & 57.5 & 69.0 \\
\midrule
\multirow{4}{*}{CFQ}          & Random & 0.0 & 4.2 & 37.6 & 91.7 & 97.9 & 15.3 & 33.2 & 55.0 & 35.4 & 51.8 & 76.5 \\
                              & MCD1 & 0.0 & 2.1 & 14.6 & 42.9 & 66.9 & 1.0 & 4.3 & 10.5 & 8.6 & 16.8 & 29.4 \\
                              & MCD2 & 0.0 & 0.0 & 2.0 & 21.4 & 27.8 & 0.6 & 2.8 & 5.3 & 1.7 & 9.6 & 15.1 \\
                              & MCD3 & 0.0 & 0.0 & 12.4 & 14.9 & 19.6 & 1.3 & 5.7 & 9.5 & 4.2 & 15.1 & 22.7 \\
\midrule      
\multirow{8}{*}{GeoQuery}     & Std. & 73.6 & 85.0 & 91.8 & 92.9 & 93.6 & 63.2                               & 81.4 & 86.8 & 77.5 & 89.6 & 92.1 \\
                              & Template1 & 72.6 & 80.4 & 78.3 & 85.8 & 89.5 & 25.9 & 64.5 & 74.1 & 50.6 & 71.1 & 83.7 \\
                              & Template2 & 73.1 & 79.5 & 85.5 & 90.0 & 91.2 & 37.5 & 69.2 & 81.6 & 54.1 & 68.9 & 68.0 \\
                              & Template3 & 60.0 & 68.5 & 74.2 & 79.4 & 82.4 & 26.7 & 58.5 & 74.2 & 41.5 & 73.6 & 82.1 \\
                              & TMCD1 & 57.9 & 59.7 & 66.4 & 67.3 & 83.3 & 39.1 & 51.2 & 60.9 & 46.7 & 57.3 & 67.0 \\
                              & TMCD2 & 48.8 & 60.6 & 61.5 & 65.2 & 73.6 & 44.5 & 49.1 & 60.0 & 36.7 & 61.8 & 73.9 \\
                              & TMCD3 & 58.2 & 58.5 & 66.4 & 54.5 & 86.7 & 43.9 & 57.6 & 70.0 & 59.1 & 67.3 & 80.6 \\
                              & Length & 32.1 & 30.9 & 33.6 & 40.3 & 41.5 & 28.8 & 40.6 & 57.9 & 31.5 & 52.4 & 63.9 \\
\midrule
\multirow{7}{*}{SMCalFlow-CS} & 8-S & 9.4 & 31.9 & 58.9 & 73.9 & 83.1 & 31.7 & 47.5 & 58.3 & 68.6 & 80.6 & 85.6 \\
                              & 8-C & 0.0 & 0.0 & 0.0 & 0.0 & 0.0 & 0.0 & 0.8 & 4.7 & 9.7 & 10.8 & 33.9 \\
                              & 16-S & 7.8 & 30.6 & 62.5 & 72.8 & 82.8 & 31.1 & 47.8 & 60.0 & 68.6 & 80.6 & 85.6 \\
                              & 16-C & 0.0 & 0.0 & 0.0 & 0.0 & 10.0 & 1.1 & 2.2 & 5.0 & 15.8 & 19.4 & 36.7 \\
                              & 32-S & 15.0 & 30.3 & 63.6 & 73.9 & 82.5 & 32.5 & 48.3 & 59.2 & 68.6 & 80.6 & 85.6 \\
                              & 32-C & 0.0 & 0.0 & 0.0 & 0.0 & 23.6 & 2.5 & 6.4 & 11.7 & 26.9 & 32.5 & 45.6 \\
                              & Length & 1.7 & 4.4 & 29.4 & 38.1 & 59.4 & 3.3 & 8.1 & 13.9 & 22.5 & 39.4 & 46.9 \\
\bottomrule
\end{tabular}}
\caption{Prompt tuning (PT) and in-context learning (ICL) results on all datasets and splits.}
\label{tab:pt_icl_results}
\end{table*}

\subsection{Example Prediction Errors}
We show example prediction errors of fine-tuned T5 models in Table~\ref{tab:errors}.

\begin{table*}[t]
\begin{center}
\scalebox{0.8}{
\begin{tabular}{ll}
\toprule
\emph{\bf Source:} & how long is the longest river in the m0 \\
\emph{\bf Target:} & \texttt{answer ( len ( longest ( intersection ( river , loc\_2 ( m0 ) ) ) ) )} \\
\emph{\bf T5-small Prediction:} & \texttt{answer ( \textcolor{red}{longest} ( intersection ( river , loc\_2 ( m0 ) ) ) ) \textcolor{red}{)}} \\
\emph{\bf T5-base Prediction:} & \texttt{answer ( \textcolor{red}{longest} ( intersection ( river , loc\_2 ( m0 ) ) ) )} \\
\emph{\bf T5-large Prediction:} & \texttt{answer ( len ( longest ( intersection ( river , loc\_2 ( m0 ) ) ) ) )} \\
\emph{\bf T5-3B Prediction:} & \texttt{answer ( len ( longest ( intersection ( river , loc\_2 ( m0 ) ) ) ) )} \\
\emph{\bf T5-11B Prediction:} & \texttt{answer ( len ( longest ( intersection ( river , loc\_2 ( m0 ) ) ) ) )} \\
\\

\emph{\bf Source:} & make an event with my manager \\
\emph{\bf Target:} & \texttt {(Yield (CreateCommitEventWrapper (CreatePreflightEventWrapper } \\
                   & \myquad[1] \texttt{(Event.attendees\_? (AttendeeListHasRecipient } \\
                   & \myquad[2] \texttt{(FindManager (toRecipient (CurrentUser)))))))) } \\
\emph{\bf T5-small Prediction:} & \texttt {(Yield (CreateCommitEventWrapper (CreatePreflightEventWrapper } \\
                   & \myquad[1] \texttt{(Event.attendees\_? (AttendeeListHasRecipient } \\
                   & \myquad[2] \texttt{(FindManager (toRecipient (CurrentUser\textcolor{red}{)))))} } \\
\emph{\bf T5-base Prediction:} & \texttt {(Yield (CreateCommitEventWrapper (CreatePreflightEventWrapper } \\
                   & \myquad[1] \texttt{(Event.attendees\_? (AttendeeListHasRecipient } \\
                   & \myquad[2] \texttt{(FindManager (toRecipient (CurrentUser\textcolor{red}{))))))} } \\
\emph{\bf T5-large Prediction:} &  \texttt {(Yield (CreateCommitEventWrapper (CreatePreflightEventWrapper } \\
                   & \myquad[1] \texttt{(Event.attendees\_? (AttendeeListHasRecipient } \\
                   & \myquad[2] \texttt{(FindManager (toRecipient (CurrentUser\textcolor{red}{)))))))} } \\
\emph{\bf T5-3B Prediction:} & \texttt {(Yield (CreateCommitEventWrapper (CreatePreflightEventWrapper } \\
                   & \myquad[1] \texttt{\textcolor{red}{(Event.subject\_? (?= "event with my manager"))}))) } \\
\emph{\bf T5-11B Prediction:} & \texttt {(Yield (CreateCommitEventWrapper (CreatePreflightEventWrapper } \\
                   & \myquad[1] \texttt{(Event.attendees\_? (AttendeeListHasRecipient } \\
                   & \myquad[2] \texttt{(FindManager (toRecipient (CurrentUser)))))))) } \\

\midrule

\emph{\bf Source:} & which states have cities named m0 \\
\emph{\bf Target:} & \texttt{answer ( intersection ( state , loc\_1 ( intersection ( city , m0 ) )  ) )} \\
\emph{\bf T5-small Prediction:} & \texttt{answer ( intersection ( state , loc\_1 ( intersection ( city , m0 ) ) ) )} \\
\emph{\bf T5-base Prediction:} & \texttt{answer ( intersection ( state , loc\_1 ( intersection (   city , \textcolor{red}{loc\_2 ( m0 )} ) ) ) )} \\
\emph{\bf T5-large Prediction:} & \texttt{answer ( intersection ( state , \textcolor{red}{loc\_2 ( m0 )} ) )} \\
\emph{\bf T5-3B Prediction:} & \texttt{answer ( intersection ( state , loc\_1 ( \textcolor{red}{m0} ) ) )} \\
\emph{\bf T5-11B Prediction:} & \texttt{answer ( intersection ( state , \textcolor{red}{loc\_2 ( m0 )} ) )} \\
\\

\emph{\bf Source:} & Make me an event for 3 pm tomorrow with my team \\
\emph{\bf Target:} & \texttt{(Yield (CreateCommitEventWrapper (CreatePreflightEventWrapper } \\
                   & \myquad[1] \texttt{(\& (Event.start\_? (?= (DateAtTimeWithDefaults (Tomorrow) (NumberPM 3L)))) } \\
                   & \myquad[2] \texttt{(Event.attendees\_? (AttendeeListHasPeople } \\
                   & \myquad[3] \texttt{(FindTeamOf (toRecipient (CurrentUser)))))))))} \\
\emph{\bf T5-small Prediction:} & \texttt{(Yield (CreateCommitEventWrapper (CreatePreflightEventWrapper } \\
                   & \myquad[1] \texttt{(\& (Event.start\_? (?= (DateAtTimeWithDefaults (Tomorrow) (NumberPM 3L)))) } \\
                   & \myquad[2] \texttt{(Event.attendees\_? (AttendeeListHasPeople } \\
                   & \myquad[3] \texttt{(FindTeamOf (toRecipient (CurrentUser)))))))))} \\
\emph{\bf T5-base Prediction:} & \texttt{(Yield (CreateCommitEventWrapper (CreatePreflightEventWrapper } \\
                   & \myquad[1] \texttt{(\& (Event.start\_? (?= (DateAtTimeWithDefaults (Tomorrow) (NumberPM 3L)))) } \\
                   & \myquad[2] \texttt{(Event.attendees\_? (AttendeeListHasPeople } \\
                   & \myquad[3] \texttt{(FindTeamOf (toRecipient (CurrentUser)))))))))} \\
\emph{\bf T5-large Prediction:} & \texttt{(Yield (CreateCommitEventWrapper (CreatePreflightEventWrapper } \\
                   & \myquad[1] \texttt{(\& \textcolor{red}{(Event.subject\_? (?= "event with my team"))} } \\
                   & \myquad[2] \texttt{(Event.start\_? (?= (DateAtTimeWithDefaults (Tomorrow) (NumberPM 3L))))))))} \\
\emph{\bf T5-3B Prediction:} & \texttt{(Yield (CreateCommitEventWrapper (CreatePreflightEventWrapper } \\
                   & \myquad[1] \texttt{(\& \textcolor{red}{(Event.subject\_? (?= "my team"))} } \\
                   & \myquad[2] \texttt{(Event.start\_? (?= (DateAtTimeWithDefaults (Tomorrow) (NumberPM 3L))))))))} \\
\emph{\bf T5-11B Prediction:} & \texttt{(Yield (CreateCommitEventWrapper (CreatePreflightEventWrapper } \\
                   & \myquad[1] \texttt{(\& \textcolor{red}{(Event.subject\_? (?= "my team"))} } \\
                   & \myquad[2] \texttt{(Event.start\_? (?= (DateAtTimeWithDefaults (Tomorrow) (NumberPM 3L))))))))} \\
\bottomrule
\end{tabular}
}
\caption{Example predictions of fine-tuned T5 models on the development set of GeoQuery and SMCalFlow-CS dataset where the errors are corrected (top) and caused (bottom) by model scale.}
\label{tab:errors}
\end{center}
\end{table*}

\end{document}